\documentclass[letterpaper]{article} 
\usepackage[]{aaai2027}  
\usepackage[hyphens]{url}  
\usepackage{graphicx} 
\usepackage{natbib}  
\usepackage{caption} 
\usepackage{algorithm}
\usepackage{algorithmic}
\usepackage{xspace}
\usepackage{enumitem}

\usepackage{newfloat}
\usepackage{listings}
\DeclareCaptionStyle{ruled}{labelfont=normalfont,labelsep=colon,strut=off} 
\floatstyle{ruled}
\newfloat{listing}{tb}{lst}{}
\floatname{listing}{Listing}

\newcommand{\method}{\texttt{ZIPBrain}\xspace }
\usepackage{booktabs}

\usepackage{amsmath}
\usepackage{amssymb}
\usepackage{amsthm}

\usepackage{threeparttable}
\usepackage{multirow}
\usepackage{makecell}

\usepackage{subcaption}

\DeclareMathOperator*{\argmax}{argmax}
\title{ZIPBrain: Can EEG Foundation Models Be Faster, Locally Deployable, \\but Accurate?}
\author {
    Lingwei Li\textsuperscript{\rm 1},
    Yirong Kan\textsuperscript{\rm 1},
    Peng Chen\textsuperscript{\rm 2},
    Xu Cao\textsuperscript{\rm 3},
    Zheng Chen\textsuperscript{\rm 4},
    Yasuhiko Nakashima\textsuperscript{\rm 1}
}
\affiliations {
    \textsuperscript{\rm 1}Nara Institute of Science and Technology, Nara, Japan\\
    \textsuperscript{\rm 2}RIKEN Center for Computational Science, Hyogo, Japan\\
    \textsuperscript{\rm 3}University of Illinois Urbana-Champaign, USA\\
    \textsuperscript{\rm 4}SANKEN, The University of Osaka, Japan\\
    li.lingwei.lo2@naist.ac.jp, kan.yirong@is.naist.jp, peng.chen@riken.jp, xucao2@illinois.edu, chenz@sanken.osaka-u.ac.jp, nakashim@is.naist.jp
    
}

\nocopyright

\begin{document}

\maketitle

\begin{abstract}

This work investigates whether Electroencephalograph (EEG) foundation models (EFMs) can be made faster and locally deployable without sacrificing accuracy. EEG foundation models are a major trend, offering strong general-purpose representations. However, their computational burden grows quadratically with input length, hindering deployment on resource-constrained scenario, particularly for real-time clinical monitoring. 
EEG's low SNR further suggests many of these tokens are redundant and compressible with little accuracy cost. We propose ZIPBrain, a novel redundancy-aware EEG token pooling module that leverages this low-SNR characteristic to reduce token count. Given a token sequence, ZIPBrain partitions tokens into redundant and unique groups, then merges each redundant token with its most similar counterpart in the unique group. Furthermore, ZIPBrain serves as a training-free, plug-and-play module that seamlessly integrates into standard Transformer encoders with negligible computational overhead. Extensive experiments across multiple EEG foundation models show ZIPBrain's strong versatility, achieving 1.3\%-10.5\% average improvement over baselines, while reducing wall-clock inference time by 32.7\% (up to 41.8\% with CUDA Graph) compared to the original foundation model.

\end{abstract}

\section{Introduction}
\label{sec:intro}

The breakthroughs in Large Language Models (LLMs)~\cite{radford2019language, brown_language_2020, openai2024gpt4technicalreport, team_gemini_2025} have driven a broader shift toward scaling attention-based architectures, giving rise to a series of EEG foundation models that push the state of the art on downstream tasks such as seizure detection, Brain-Computer Interfaces (BCIs), and sleep staging \citep{ma_codebrain_2025,pradeepkumar_tokenizing_2025,yang2026areiclr}. 
Transformers~\cite{vaswani_attention_2017, devlin_bert_2019}, which are backbones of LLMs, require discrete token sequences as input. 
EEG signals are therefore segmented into patches, which are then embedded into tokens; self-attention operates over these tokens to capture dependencies across time and channels.

\begin{figure}[ht]
    \centering
    \includegraphics[scale=1]{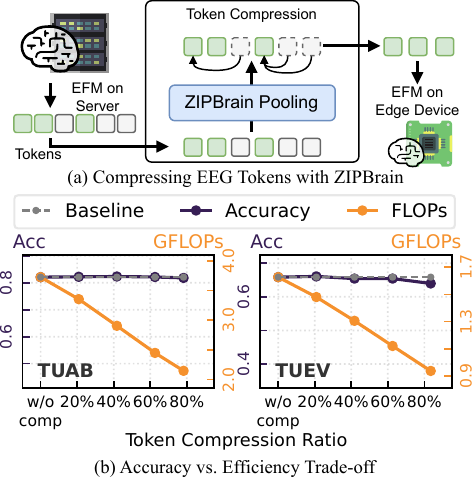}
    \caption{(a) The proposed method ZIPBrain can compress tokens of EFM on server and deploy EFM efficiently on local devices. (b) Our method achieved 42\% reduction of computations with negligible impact on accuracy. 
    }
    \label{fig:story}
\end{figure}

However, efficiency remains a critical bottleneck: computing pairwise correlations across all tokens gives Transformers quadratic complexity in sequence length, both in computation and deployment. 
This burden is particularly acute for EEG, where long-duration or high-density recordings can easily yield token sequences of several thousand in length, restricting deployment in resource-constrained or real-time clinical monitoring scenarios. 
The problem is compounded by the low signal-to-noise ratio (SNR) intrinsic to EEG data \citep{ AAAI2025SODOR}: genuine neural activity is often buried beneath physiological artifacts and background noise, so a substantial portion of the resulting tokens carries largely non-discriminative information rather than task-relevant signal.

A few lines of research have been proposed to mitigate efficiency bottleneck and redundancy issue. Some studies propose to reduce the token count right at the data preprocessing stage by selectively sampling patches \citep{ankireddy_timesqueeze_2025, zhou_adaptive_2026}. Another line bypasses the quadratic complexity entirely by substituting the Transformer architecture with SSMs \citep{tegon_femba_2025}. 
However, the objective of the above works is data preprocessing rather than token reduction.
In this work, we investigate whether token compression can be applied to EEG foundation models to mitigate efficiency bottleneck by leveraging the low-SNR property of EEG signals.
Token compression have been studied in Computer Vision domain, including importance-based compression and redundancy-based compression.
Importance-based token compression \cite{liang_evit_2021, zhang_textvisual_2025} focus on identifying informative tokens to preserve, contrastively, emerging in recent years, redundancy-based token compression \cite{wen_stop_2025, fang_prune_2025} focus on reducing tokens containing redundancy information.
However, we acknowledge that leveraging these methods to EEG foundation models is non-trivial and largely unexplored; and we still face three major challenges:
(1) Token Selection. EEG tokens carry no explicit visual or semantic cues. Unlike CV or NLP, there is no clear signal for which tokens matter most. 
(2) Redundant Token Identification. EEG signals have a low signal-to-noise ratio. Many tokens are therefore redundant, not task-irrelevant. Quantifying this redundancy remains largely unexplored. 
(3) Merging for Compression. Naively averaging redundant tokens with their target counterparts dampens feature magnitude. An averaged vector's norm is bounded by its constituents' maximum. 
This risks losing salient EEG signatures.

To tackle these challenges, we present \method (as shown in Figure 1(a)), a plug-and-play, training-free token pooling framework for EEG foundation models. Specifically, we introduce an $l$2-norm-based pivot selection mechanism to identify a representative token set without relying on CV- or NLP-specific priors; define an explicit redundancy score, computed as each token's accumulated cosine similarity to the selected pivots, to partition tokens into unique and redundant groups; and merge matched tokens through a norm-preserving pooling operation that rescales the averaged representation to the maximum constituent norm, avoiding the feature attenuation inherent to naive averaging. 
As shown in Figure \ref{fig:story}(b), \method preserved 99.65\% and 97.14\% of LaBraM's downstream accuracy on TUAB and TUEV dataset at $\sim$80\% compression ratio, with 42.30\% and 42.36\% reduction in computational burdens, and improved the accuracy by 0.2\% at $\sim$40\% compression ratio on TUAB. In empirical evaluations across four EEG foundation models and five datasts,  \method{} achieved 16 top-1 results and 20/20 top-2 results out of 20 settings compared to baselines.

\begin{figure*}[ht]
    \centering
    \includegraphics[scale=1]{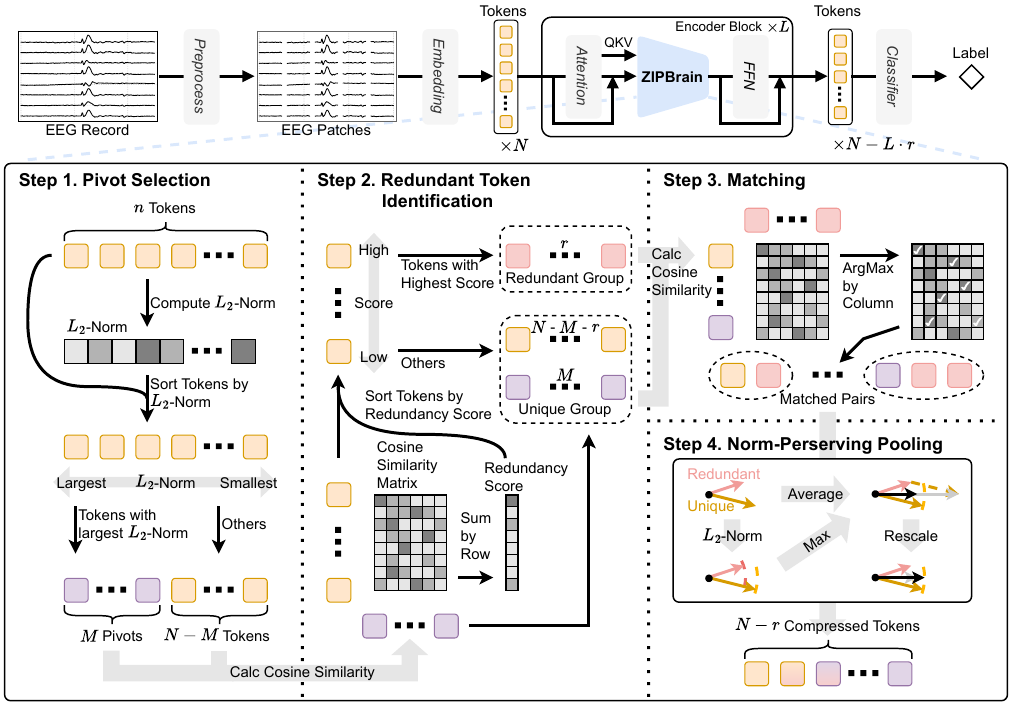}
    \caption{{Overview of the proposed module ZIPBrain.} The flow-chart in the top illustrates the integration of our method into a Transformer block, positioned between the Attention and Feed-Forward layers to leverage intermediate $Q$, $K$, or $V$ representations. The bottom panel details three processes of the method: (1) \textit{Redundancy-based partitioning}, where tokens are partitioned into unique and redundant sets based on redundancy scores calculated relative to pivots, which are selected from the input tokens by ranking their $\ell_2$-norms in descending order and assigned to unique group; (2) \textit{Similarity-based matching}, where each redundant token is matched with its most similar counterpart in unique group; and (3) \textit{Norm-preserving pooling}, which includes a rescaling operation to preserve the original norm characteristics of the tokens, following taking average of tokens.}
    \label{fig:method_overview}
\end{figure*}

\section{Related Work}
\label{sec:related}
\noindent \textbf{EEG Foundation Models.}
Trained as a foundation model, BIOT~\cite{yang_biot_2023} validated that the pre-training paradigm yields superior representations compared to supervised training methods. 
Subsequent frameworks, including LaBraM~\cite{jiang_large_2024}, EEGPT~\cite{wang_eegpt_2024}, CBraMod~\cite{wang_cbramod_2024}, TFM-Tokenizer~\cite{pradeepkumar_tokenizing_2025}, CodeBrain~\cite{ma_codebrain_2025}, and ST-EEGFormer~\cite{yang2026areiclr}, introduced diverse pre-training objectives and tokenization schemes tailored to EEG dynamics, consistently establishing new state-of-the-art benchmarks. 
However, the deployment of these large-scale models is often constrained by heavy computational overhead, forcing a strategic trade-off between model performance and computational efficiency.
\smallskip

\noindent \textbf{Token Compression.} The token compression has been widely explored in CV to reduce computational complexity by reducing the number of tokens. Existing approaches can be broadly categorized into \textit{importance-based} and \textit{redundancy-based} strategies. 
\textit{Importance-based methods} identify and reduce the number of uninformative tokens by evaluating token significance using auxiliary predictor networks~\cite{rao_dynamicvit_2021a}, [CLS] attention scores~\cite{liang_evit_2021}, or cross-modal attention maps in VLMs~\cite{chen_image_2025, zhang_textvisual_2025, ye_fit_2025, arif_hired_2025}. 
Conversely, \textit{redundancy-based methods} focus on reduce the number of redundant tokens, employing techniques such as bipartite soft matching~\cite{bolya_token_2023} or explicit redundancy metrics~\cite{fang_prune_2025, dhouib_pact_2025}.\\
\textit{\textbf{-Issues.}}
Despite its proved success in visual domains, token compression in the context of EEG Transformers remains largely under-explored. EEG signals exhibit fundamentally different characteristics from images, transferring vision-oriented compression strategies to EEG models naively may risk discarding critical neural information, resulting a retrogradation in model capability.




\section{Problem Definition}
\label{sec:prelim}



This section presents the problem formulation of token compression for EEG foundation models. 

\noindent\textbf{Definition 1 (Multi-Channel EEG Tokens).} 
EEG recordings capture neuronal activity across multiple brain regions over time.
Let $\mathbf{S} \in \mathbb{R}^{C \times T}$ denote an EEG recording, where $C$ is the number of channels and $T$ is the number of samples.

The recording $\mathbf{S}$ is segmented using a sliding window of size $w$ with stride $s$, producing
$\hat{\mathbf{S}} \in \mathbb{R}^{C \times \tau \times w}$,
where
$\tau=\left\lfloor\frac{T-w}{s}\right\rfloor+1$
is the number of temporal segments.
Each segment is then projected into a $d$-dimensional embedding, yielding the initial token sequence
$\mathbf{X}^{0}=\{\mathbf{x}_i^{0}\}_{i=1}^{N}\in\mathbb{R}^{N\times d}$,
where $\mathbf{x}_i^{0}\in\mathbb{R}^{d}$ denotes the $i$-th token,
$N=C\tau$ is the total number of tokens,
and $d$ is the embedding dimension.
The superscript $l$ denotes the Transformer layer, such that $\mathbf{X}^{0}$ represents the input tokens and $\mathbf{X}^{l}$ the token representations at the $l$-th layer.

\noindent\textbf{Definition 2 (Token Pooling).} Given a sequence of tokens, token pooling aims to merge tokens together
\begin{equation}
[\tilde{X}_1, \cdots, \tilde{X}_{n_2}]:=f_{\text{pool}}(X_1, X_2,\dots,X_{n_1}),
\end{equation}
where $n_2<n_1$ and $f_{\text{pool}}$ can be any operation that implements this behavior.

However, implementing $f_\text{pool}$ is non-trivial for the preservation of model's capability, with following three problems. 

\paragraph{Problem 1 (Token Selection).}
One of the core challenges in token merging is to determine an optimal partition of the input tokens into $N'$ groups, denoted by $\mathcal{G}=\{\mathbf{G}_1,\mathbf{G}_2,\ldots,\mathbf{G}_{N'}\}$, where $\mathbf{G}_j$ is the $j$-th token group. Given the input token sequence $\mathbf{X}\in\mathbb{R}^{N\times d}$ and the target sequence length $N'$, the grouping can be represented by a binary assignment matrix $\mathbf{M}=\{m_{ij}\}\in\{0,1\}^{N\times N'}$, where
\[
\mathbf{G}_j=\{\mathbf{x}_i\mid m_{ij}=1\}.
\]
Here, $m_{ij}=1$ indicates that $\mathbf{x}_i$ is assigned to group $j$, subject to the partition
$
\sum_{j=1}^{N'} m_{ij}=1,\qquad \forall i\in\{1,\ldots,N\},
$
ensuring that each token is assigned to one group.

\paragraph{Problem 2 (Redundancy of EEG Tokens).}
In the context of EEG processing, tokens naturally exhibit substantial redundancy. Effectively exploiting this redundancy to guide token grouping remains a fundamental challenge in token compression. Formally, given the token sequence $\mathbf{X}=\{\mathbf{x}_i\}_{i=1}^{N}$, the objective is to define a redundancy measure
$
r(\mathbf{x}_i,\mathbf{X}),
$
which quantifies the redundancy of given token $\mathbf{x_i}$ in population $\mathbf{X}$ and guides the construction of the token partition $\mathcal{G}$.

\paragraph{Problem 3 (Merging Strategy).}
Given a token group $\mathbf{G}_j$, the final challenge is to define a merging function $f_{\mathrm{merge}}$ that aggregates all tokens within the group into a single representative token,
\begin{equation}
    \hat{\mathbf{x}}_{j}
    =f_{\mathrm{merge}}\!\left(\{\mathbf{x}\mid\mathbf{x}\in\mathbf{G}_j\}\right).
\end{equation}
Various choices exist for implementing $f_{\mathrm{merge}}$, ranging from simple mean pooling to more sophisticated attention-based aggregation strategies.

\section{Method}
\label{sec:moethod}
\subsection{Overview of \method}
\label{sec:method_overview}

To tackle the above challenges, we propose \method, a plug-and-play token compression method for Transformer-based EEG foundation models. 
As illustrated in the top panel of Figure~\ref{fig:method_overview}, \method is seamlessly inserted between the self-attention module and the feed-forward network (FFN), enabling token compression without modifying the Transformer backbone. 
Given the intermediate tokens produced by the attention, \method compresses them through three sequential components, as shown in the bottom panel of Figure~\ref{fig:method_overview}: (1) a two-stage partitioning module (Step 1 and 2) that partitions tokens into two sets, with a redundancy-aware partitioning strategy that exploits the intrinsic redundancy of EEG tokens, (2) a token matching module (Step 3) that further construct groups by identifying merging target for redundant tokens, and (3) a norm-preserving merging module (Step 4) that aggregates each group into a representative token while preserving informative representations. 
Together, these components effectively reduce the sequence length while maintaining the discriminative information required for downstream EEG analysis.





\subsection{Step 1: Pivots Selection}
The two-stage partitioning module categorizes tokens into a redundant set and a unique set based on their redundancy. This module begins with a Pivot Selection stage, which selects a subset of tokens---termed pivots---to represent the collective information of the original tokens. Specifically, the pivots $\mathbf{X}_p$ are chosen as the tokens with the largest $\ell_2$-norms.
We use $\mathcal{I}_p$ to represent the indices set of pivots, define
\begin{equation}
    \mathcal{I}_p = \mathop{\arg\max}_{\mathcal{S} \subseteq \{1, \dots, N-r\}, |\mathcal{S}| = N_p} \sum_{i \in \mathcal{S}} \|\mathbf{x}_i\|_2,
\end{equation}
and define $\mathbf{X}_p$ as $\mathbf{X}_p = \mathbf{x}_{i\in\mathcal{I}_p}$ to represent the set of pivotal tokens.
The $N_p = \lceil (N - r) \cdot \rho \rceil$ denotes the number of selected pivots, and is controlled by a hyperparameter ratio $\rho \in (0,1]$ that ensures $N_p \in [1, N-r]$. 





\subsection{Step 2: Redundant Token Identification}

Since pivots capture the core contextual information, a token's redundancy naturally corresponds to its alignment with these pivots. Formally, we evaluate the redundancy score $s_i$ for each token $\mathbf{x}_i$ by aggregating its similarities across all pivots:
\begin{equation}
    s_i = 
    \begin{cases}
        0, & \text{if } i \in \mathcal{I}_p, \\
        \sum_{j \in \mathcal{I}_p} \tilde{\mathbf{x}}_i \tilde{\mathbf{x}}_j^\top, & \text{otherwise,}
    \end{cases}
\end{equation}
where $\tilde{\mathbf{x}}_i$ and $\tilde{\mathbf{x}}_j$ denote the $\ell_2$-normalized forms of $\mathbf{x}_i$ and $\mathbf{x}_j$, respectively, and $\tilde{\mathbf{x}}_i \tilde{\mathbf{x}}_j^\top$ denotes the cosine similarity between candidate token $\mathbf{x}_i$ and pivot token $\mathbf{x}_j$. Assuming row vectors $\mathbf{x}_i \in \mathbb{R}^{1 \times d}$, the matrix product $\tilde{\mathbf{x}}_i \tilde{\mathbf{x}}_j^\top$ produces a scalar. The summation $\sum_{j \in \mathcal{I}_p} \tilde{\mathbf{x}}_i \tilde{\mathbf{x}}_j^\top$ measures the cumulative cosine similarity and can be efficiently computed by associative reordering: $\tilde{\mathbf{x}}_i \left( \sum_{j \in \mathcal{I}_p} \tilde{\mathbf{x}}_j^\top \right)$. Notably, pivots are assigned a zero score to ensure their preservation during token compression.

Tokens are assigned to either the redundant set or the unique set depending on their redundancy scores. Formally, we denote the redundant and unique token sets as $\mathbf{X}_r$ and $\mathbf{X}_u$, respectively, with corresponding index sets $\mathcal{I}_r$ and $\mathcal{I}_u$, defined as $\mathbf{X}_r = \{\mathbf{x}_i \mid i \in \mathcal{I}_r\}$ and $\mathbf{X}_u = \{\mathbf{x}_i \mid i \in \mathcal{I}_u\}$.
Tokens with the top-$r$ redundancy scores are assigned to the redundant set $\mathbf{X}_r$, while the remaining tokens are assigned to the unique set $\mathbf{X}_u$. Therefore, the index sets $\mathcal{I}_r$ and $\mathcal{I}_u$ are defined as:
\begin{equation}
\begin{aligned}
    \mathcal{I}_r &= \operatorname*{argmax}_{\mathcal{S} \subseteq \{1, \dots, N\}, |\mathcal{S}| = r} \sum_{i \in \mathcal{S}} s_i, \\
    \mathcal{I}_u &= \{1, \dots, N\} \setminus \mathcal{I}_r,
\end{aligned}
\end{equation}
where $\setminus$ denotes the set difference operation.
\subsection{Step 3: Matching For Token Pairs.} 
For each redundant token $\mathbf{x}_{i\in\mathcal{I}_p}$, we match it with the most similar counterpart in unique set and use function $\mathcal{M}(i\in\mathcal{I}_r) \to j \in \mathcal{I}_u$ to formulate this projection as
\begin{equation}
    \mathcal{M}(i) = \arg\max_{j \in \mathcal{I}_u} \left( \tilde{\mathbf{x}}_{i} \tilde{\mathbf{x}}_{j}^T \right), \forall i \in \mathcal{I}_r,
    \quad
\end{equation}
where $\tilde{\mathbf{x}}_{i},\tilde{\mathbf{x}}_{j}$ denote $\ell_2$-normalized ${\mathbf{x}}_{i},{\mathbf{x}}_{j}$ and $\tilde{\mathbf{x}}_{i} \tilde{\mathbf{x}}_{j}^T$ calculated the cosine similarity between two vectors.
Groups are then defined on top of $\mathcal{M}(\cdot)$ as 
\begin{equation}
\begin{aligned}
\mathcal{G} & = \{\mathbf{G}_j, j\in[1,\dots,N']\} \\ & = \{ \{\mathbf{x}_i\} \cup \{\mathbf{x}_j | \mathcal{M}(j)=i, j\in \mathcal{I}_r\}, \forall i \in \mathcal{I}_u\},
\end{aligned}
\end{equation}
which provides the necessary pairing information for the next token merging step.

\subsection{Step 4: Norm-Preserving Feature Merging}
\label{sec:merging}

The merging mechanism operates on the principle of norm preservation across all tokens within $\mathbf{G}_j$. It first calculates the average direction of the cluster tokens and then scales this normalized vector by the maximum feature norm observed among the individual vectors in the set:
\begin{equation}
    \bar{\mathbf{x}}_j = \frac{1}{|\mathbf{G}_j|} \sum_{\mathbf{x}_k \in \mathbf{G}_j} \mathbf{x}_k,
\end{equation}
\begin{equation}
    \tilde{\mathbf{x}}_j = \frac{\bar{\mathbf{x}}_j}{\|\bar{\mathbf{x}}_j\|_2} \cdot \max_{\mathbf{x}_k \in \mathbf{G}_j} \|\mathbf{x}_k\|_2,
\end{equation}
where $\tilde{\mathbf{x}}_j \in \mathbb{R}^d$ represents the final condensed token representation that forms the $j$-th row of the compressed feature matrix $\tilde{\mathbf{X}} \in \mathbb{R}^{N' \times d}$. 

\begin{table*}[t]
    \centering
    \small
    \begin{threeparttable}
\newcolumntype{H}{>{\setbox0=\hbox\bgroup}c<{\egroup}@{}}
\begin{tabular}{ll Hcc Hcc | Hcc Hcc Hcc}
    \toprule
    \multirow{2}{*}{Model} & \multirow{2}{*}{Methods} & \multicolumn{3}{c}{TUAB} & \multicolumn{3}{c}{EEGMAT} & \multicolumn{3}{c}{TUEV} & \multicolumn{3}{c}{ISRUC} & \multicolumn{3}{c}{EarEEG} \\
    \cmidrule(lr){3-5} \cmidrule(lr){6-8} \cmidrule(lr){9-11} \cmidrule(lr){12-14} \cmidrule(lr){15-17}
    & & Hyper-Params & \small AUROC & \small PR-AUC & Hyper-Params & \small AUROC & \small PR-AUC & Hyper-Params & Kappa\footnotemark[2] & W-F1\footnotemark[2] & Hyper-Params & Kappa & W-F1 & Hyper-Params & Kappa & W-F1 \\

    \midrule
    \multirow{6.5}{*}{\small LaBraM} 
    & Original\footnotemark[1] &  & 0.9067 & 0.9092 &  & 0.6230 & 0.6244 &  & 0.6616 & 0.8292 &  & 0.7383 & 0.7958 &  & 0.4255 & 0.5413 \\
    \cmidrule(lr){2-17}
    & EVIT &  & 0.8529 & 0.8501 &  & 0.6342 & 0.6316 &  & 0.5766 & 0.7922 &  & 0.5972 & 0.6849 &  & 0.2474 & 0.4003 \\
    & ToMe &  & 0.8854 & 0.8893 &  & 0.6086 & 0.6042 &  & 0.6227 & 0.8106 &  & 0.6084 & 0.6951 &  & 0.3072 & 0.4476 \\
    & ToFU &  & 0.8884 & \textbf{0.8928} &  & 0.6154 & 0.6234 &  & 0.6272 & 0.8129 &  & 0.6332 & 0.7139 &  & 0.3069 & 0.4497 \\
    & DART &  & 0.8731 & 0.8713 &  & 0.6207 & 0.6077 &  & 0.5416 & 0.7704 &  & 0.6642 & 0.7348 &  & 0.3489 & 0.4805 \\
    & Ours & x,v,4,x,10 & \textbf{0.8906} & \underbar{0.8916} & x,q,3,x,10 & \textbf{0.6477} & \textbf{0.6751} & k,k,9,k,10 & \textbf{0.6479} & \textbf{0.8226} & q,v,1,x,10 & \textbf{0.6763} & \textbf{0.7508} & x,vh,1,kh,10 & \textbf{0.4249} & \textbf{0.5318} \\
    
    \midrule
    \multirow{6.5}{*}{EEGPT} 
    & Original\footnotemark[1] &  & 0.8684 & 0.8470 &  & 0.6517 & 0.6211 &  & 0.5677 & 0.7887 &  & 0.6587 & 0.7252 &  & 0.4797 & 0.5767 \\
    \cmidrule(lr){2-17}
    & EVIT\textsuperscript{mean} &  & 0.8665 & 0.8449 &  & 0.6512 & 0.6211 &  & 0.5603 & 0.7861 &  & 0.6544 & 0.7203 &  & 0.4558 & 0.5565 \\
    & ToMe &  & 0.8672 & 0.8457 &  & 0.6444 & 0.6108 &  & 0.5557 & 0.7850 &  & 0.6411 & 0.7111 &  & 0.4539 & 0.5543 \\
    & ToFU &  & 0.8672 & 0.8458 &  & 0.6459 & 0.6120 &  & 0.5566 & 0.7854 &  & 0.6441 & 0.7134 &  & 0.4539 & 0.5550 \\
    & DART &  & 0.8687 & 0.8468 &  & 0.6528 & 0.6201 &  & 0.5620 & 0.7872 &  & \textbf{0.6637} & \textbf{0.7301} &  & 0.4647 & 0.5634 \\
    & Ours & kh,x,0,x,10 & \textbf{0.8690} & \textbf{0.8482} & x,v,0,v,10 & \textbf{0.6563} & \textbf{0.6257} & q,q,10,q,10 & \textbf{0.5746} & \textbf{0.7889 }& x,q,5,qh,10 & \underbar{0.6612} & \underbar{0.7273} & vh,qh,6,v,10 & \textbf{0.4689} & \textbf{0.5656} \\

    \midrule
    \multirow{6.5}{*}{BIOT} 
    & Original\footnotemark[1] &  & 0.8782 & 0.8797 &  & 0.7721 & 0.7805 &  & 0.5082 & 0.7444 &  & 0.7536 & 0.8050 &  & 0.3935 & 0.5267 \\
    \cmidrule(lr){2-17}
    & EVIT\textsuperscript{mean} &  & 0.8647 & 0.8673 &  & 0.8251 & \textbf{0.8266} &  & 0.4649 & 0.7151 &  & 0.5702 & 0.6643 &  & \textbf{0.4965} & \textbf{0.5943} \\
    & ToMe &  & 0.8767 & 0.8765 &  & 0.7454 & 0.7425 &  & 0.5000 & 0.7412 &  & 0.7334 & 0.7905 &  & 0.4172 & 0.5325 \\
    & ToFU &  & 0.8771 & 0.8771 &  & 0.7410 & 0.7419 &  & 0.5020 & 0.7423 &  & 0.7343 & 0.7892 &  & 0.4101 & 0.5289 \\
    & DART &  & 0.8639 & 0.8660 &  & 0.7255 & 0.7309 &  & 0.4587 & 0.7138 &  & 0.6238 & 0.7180 &  & 0.2705 & 0.4042 \\
    & Ours & v,k,9,kh,10 & \textbf{0.8797} & \textbf{0.8791} & v,k,2,k,10 & \textbf{0.8286} & \underbar{0.8149} & kh,kh,9,qh,10 & \textbf{0.5199} & \textbf{0.7488} & x,kh,5,v,10 & \textbf{0.7345} & \textbf{0.7933} & k,k,0,k,10 & \underbar{0.4663} & \underbar{0.5642} \\

    \midrule
    \multirow{6.5}{*}{TFM} 
    & Original\footnotemark[1] &  & 0.8895 & 0.8964 &  & 0.6299 & 0.6210 &  & 0.5611 & 0.7713 &  & 0.7094 & 0.7684 &  & 0.4077 & 0.5250 \\
    \cmidrule(lr){2-17}
    & EVIT &  & 0.8708 & 0.8779 &  & 0.6075 & 0.5713 &  & 0.4411 & 0.7155 &  & 0.6727 & 0.7429 &  & 0.4436 & 0.5488 \\
    & ToMe &  & 0.8793 & 0.8834 &  & 0.6363 & 0.6279 &  & 0.5421 & 0.7615 &  & 0.6879 & 0.7487 &  & 0.3927 & 0.5071 \\
    & ToFU &  & 0.8793 & 0.8846 &  & 0.6435 & 0.6526 &  & 0.5399 & 0.7603 &  & 0.6921 & 0.7530 &  & 0.3862 & 0.5016 \\
    & DART &  & 0.8624 & 0.8662 &  & 0.5990 & 0.5889 &  & 0.5240 & 0.7523 &  & 0.6870 & 0.7537 &  & 0.2687 & 0.4227 \\
    & Ours & qh,qh,10,qh,10 & \textbf{0.8812} & \textbf{0.8888} & q,v,1,x,10 & \textbf{0.6727} & \textbf{0.6662} & k,x,8,vh,10 & \textbf{0.5628} & \textbf{0.7712} & x,q,5,x,10 & \textbf{0.7120} & \textbf{0.7641} & qh,vh,10,kh,10 & \textbf{0.4839} & \textbf{0.5771} \\

    \bottomrule

\end{tabular}

\begin{tablenotes}
\footnotesize 
    \item Experiment results with \(r\), the number of tokens to be reduced per reduction, set to the biggest integer smaller than \(\frac{N}{l}\). \(N\) is the number of total tokens and \(l\) is set to 4 for BIOT, TFM and EEGPT and 12 for LaBraM. For BIOT and TFM, which consist of 4 layers of encoder, and LaBraM, which consists of 12 layers of encoder, we insert reductions in every layer. For EEGPT, which consists of 8 layer of encoder, reductions are insert at the 2nd, 4th, 6th and 8th layer. All hyper-parameters are set based on recommendations from original works. Compression ratio is defined as $(N - l\cdot r) / N$.
    \item[1] We include experiment results without token compression as Original.
    \item[2] Kappa here stands for Cohen's Kappa and W-F1 stands for Weighted F1.
\end{tablenotes}

\end{threeparttable}

    \caption{Comparison on accuracy with highest compression ratio available}
    \label{table:exp_main_1r4m}
\end{table*}

By adopting this norm-preserving pooling strategy, our framework avoids the erosion of feature magnitude that inherently occurs with standard linear averaging, where the norm of the average vector is strictly bounded by the maximum norm of its constituents (i.e., $\|\bar{\mathbf{x}}_j\|_2 \le \max_{\mathbf{x}_k \in \mathcal{X}_j} \|\mathbf{x}_k\|_2$). This design ensures that the fused tokens maintain a feature energy consistent with the original latent distribution, thereby preserving the representational strength of critical EEG signatures throughout the compression sequence.

\subsection{Representation Space Selection}

The proposed \method is representation-agnostic and can operate on different latent representations within the Transformer block. Specifically, the three stages of \method—pivot selection, redundancy estimation, and token grouping—can be performed using either the post-attention representation $\hat{\mathbf{X}}$ or the Query, Key, and Value projections $\{\mathbf{Q},\mathbf{K},\mathbf{V}\}$.
Different representation spaces emphasize different characteristics of the token interactions and therefore lead to different grouping behaviors. Rather than fixing a particular representation, \method treats the representation space as a modular design choice. The optimal representation is selected through hyperparameter optimization according to the backbone architecture and downstream task.

\section{Experiments and Results}
    \label{sec:exp}
    \subsection{Experimental Setup}\label{subsec:data}

To evaluate versatility, we integrate our inference-stage, plug-and-play module into four representative EEG foundation models spanning diverse architectures and pretraining objectives: {LaBraM}~\cite{jiang_large_2024}, {EEGPT}~\cite{wang_eegpt_2024}, {BIOT}~\cite{yang_biot_2023}, and {TFM-Tokenizer}~\cite{pradeepkumar_tokenizing_2025}. Since official checkpoints are largely unavailable, we independently fine-tune all models on downstream tasks. Experiments are conducted on GTX 5090 and GTX 4090 with Ubuntu 2204. 

We compare against four CV token compression methods: {ToMe}~\cite{bolya_token_2023} (bipartite-matching token merging; official code), {ToFU}~\cite{kim_token_2024} (norm-preserving MLERP merging; re-implemented on ToMe), {EViT}~\cite{liang_evit_2021} ([CLS]-attention-guided pruning; implemented per paper), and {DART}~\cite{wen_stop_2025} (pivot-based duplicity pruning; modified for controllable compression ratios).

\begin{table}[ht]
    \centering
    \small
    \setlength{\tabcolsep}{2pt}
        \begin{tabular}{l l c c c l c}
        \toprule
        Dataset & \multirow{2}{*}{\makecell{\# of \\ Class}} & \multirow{2}{*}{ \makecell{Total \\ Samples} } & \multicolumn{2}{c}{Data Proportion} & \multirow{2}{*}{\makecell{ Bal?\textsuperscript{1} }} & \multirow{2}{*}{\makecell{ Main \\ Metric }} \\
        \cmidrule(lr){4-5}
        & & & Max & Min & \\
        \midrule
        TUAB   & Binary     & 36945  & 53.88\% & 46.12\% & $\checkmark$   & AUROC\\
        EEGMAT & Binary     & 290    & 50.00\% & 50.00\% & $\checkmark$   & AUROC\\
        TUEV   & 6          & 29421  & 66.78\% &  1.12\% & $\times$       & Kappa\textsuperscript{2}\\
        ISRUC  & 5          & 8700   & 35.86\% & 11.85\% & $\times$       & Kappa\textsuperscript{2}\\
        EarEEG & 6          & 2944   & 40.05\% &  6.93\% & $\times$       & Kappa\textsuperscript{2}\\
        \bottomrule
    \end{tabular}

    \caption{A summary of datasets and main metric used for each dataset. Bal?\textsuperscript{1} shows whether the distribution of labels are balanced or imbalanced; Kappa\textsuperscript{2} stands for Cohen's Kappa}
\end{table}
\begin{table*}[t]
    \centering
    \small
    \begin{threeparttable}
\begin{tabular}{cl cccccccc}
    \toprule
    \multirow{2.5}{*}{\textbf{Dataset}} & \multirow{2.5}{*}{\textbf{Method}} & \multicolumn{2}{c}{\textbf{20\%}} & \multicolumn{2}{c}{\textbf{40\%}} & \multicolumn{2}{c}{\textbf{60\%}} & \multicolumn{2}{c}{\textbf{80\%}} \\
    \cmidrule(lr){3-4} \cmidrule(lr){5-6} \cmidrule(lr){7-8} \cmidrule(lr){9-10}
    & & \small \textbf{AUROC} & \small \textbf{PR-AUC} & \small \textbf{AUROC} & \small \textbf{PR-AUC} & \small \textbf{AUROC} & \small \textbf{PR-AUC} & \small \textbf{AUROC} & \small \textbf{PR-AUC} \\

    \midrule
    \multirow{7}{*}{\makecell[c]{\textbf{TUAB} \\ (BIOT)}} 
    & Original\footnotemark[1]   & 0.8782 & 0.8797 & 0.8782 & 0.8797 & 0.8782 & 0.8797 & 0.8782 & 0.8797 \\
    \cmidrule{2-10}
    & ToMe & 0.8786 & 0.8789 & 0.8786 & 0.8776 & 0.8779 & 0.8760 & 0.8773 & 0.8763 \\
    & ToFU & 0.8787 & 0.8790 & 0.8787 & 0.8777 & 0.8780 & 0.8762 & 0.8775 & 0.8767 \\
    & EViT & 0.8773 & 0.8799 & 0.8754 & 0.8789 & 0.8726 & 0.8766 & 0.8691 & 0.8729 \\
    & DART & 0.8781 & 0.8802 & 0.8773 & 0.8799 & 0.8747 & 0.8777 & 0.8680 & 0.8712 \\
    & Ours & \textbf{0.8792} & \textbf{0.8805} & \textbf{0.8801} & \textbf{0.8811} & \textbf{0.8810} & \textbf{0.8812} & \textbf{0.8812} & \textbf{0.8805}\\

    \midrule
    \midrule

    \multirow{2.5}{*}{\textbf{Dataset}} & \multirow{2.5}{*}{\textbf{Method}} & \multicolumn{2}{c}{\textbf{20\%}} & \multicolumn{2}{c}{\textbf{40\%}} & \multicolumn{2}{c}{\textbf{60\%}} & \multicolumn{2}{c}{\textbf{80\%}} \\
    \cmidrule(lr){3-4} \cmidrule(lr){5-6} \cmidrule(lr){7-8} \cmidrule(lr){9-10}
    & & \textbf{Kappa\footnotemark[2]} & \textbf{W-F1\footnotemark[2]} & \textbf{Kappa} & \textbf{W-F1} & \textbf{Kappa} & \textbf{W-F1} & \textbf{Kappa} & \textbf{W-F1} \\

    \midrule
    \multirow{7}{*}{\makecell[c]{\textbf{TUEV} \\ (LaBraM)}} 
    & Original\footnotemark[1]   & 0.6616 & 0.8292 & 0.6616 & 0.8292 & 0.6616 & 0.8292 & 0.6616 & 0.8292 \\
    \cmidrule{2-10}
    & ToMe & 0.6526 & 0.8254 & 0.6493 & 0.8230 & 0.6302 & 0.8141 & 0.6290 & 0.8133 \\
    & ToFU & 0.6526 & 0.8254 & 0.6548 & 0.8251 & 0.6370 & 0.8165 & 0.6397 & 0.8195 \\
    & EViT & 0.6664 & 0.8321 & 0.6633 & 0.8308 & 0.6564 & 0.8275 & 0.6240 & 0.8123 \\
    & DART & 0.6525 & 0.8248 & 0.6338 & 0.8152 & 0.6037 & 0.8003 & 0.5806 & 0.7886 \\
    & Ours & \textbf{0.6774} & \textbf{0.8379} & \textbf{0.6735} & \textbf{0.8352} & \textbf{0.6625} & \textbf{0.8284} & \textbf{0.6539} & \textbf{0.8258} \\
   
\bottomrule
\end{tabular}

\begin{tablenotes}
    \footnotesize
    \item[1] We show results without token compression for easier comparison.
    \item[2] Kappa stands for Cohen's Kappa and W-F1 stands for Weighted-F1
\end{tablenotes}

\end{threeparttable}

    \caption{Comparison of accuracy among five token compression methods. This table shows the performance of methods across different compression configurations.}
    \label{table:exp_main_6r1m}
\end{table*}

We conduct experiments across five diverse EEG datasets: {TUAB}~\cite{obeid_temple_2016}: binary normal-vs-abnormal classification (standard 10-20 system); {EEGMAT}~\cite{zyma_electroencephalograms_2019}: binary resting-vs-arithmetic state classification; {TUEV}~\cite{obeid_temple_2016}: 6-class seizure type detection; {ISRUC-Sleep}~\cite{khalighi_isrucsleep_2016}: 5-stage PSG sleep scoring; and {EarEEG}~\cite{bjarkemikkelsen_eareeg_2025}: 6-class classification on sparse ear-centered electrodes.
Following Table~3, for binary tasks ({TUAB}, {EEGMAT}), we report Accuracy, Balanced Accuracy, PR-AUC, and {AUROC} (primary metric). For multi-class tasks with severe class imbalance ({TUEV}, {ISRUC}, {EarEEG}), we report Accuracy, Balanced Accuracy, Weighted-F1, and {Cohen's Kappa} (primary metric, robust to chance agreement and skew).

\subsection{Main Results on Accuracy}
\label{subsec:main_results}

\paragraph{Performance under Maximum Compression.}
Table~\ref{table:exp_main_1r4m} reports the performance of four EEG foundation models on five datasets under the maximum available compression ratio for each architecture. \method consistently outperforms all existing token reduction methods (EViT, ToMe, ToFu, and DART) across almost all settings. While competing approaches exhibit substantial performance degradation under extreme compression, \method preserves task-relevant tokens and achieves the best accuracy--efficiency trade-off.

\paragraph{Robustness Across Varying Compression Ratios.}
Table~\ref{table:exp_main_6r1m} presents the performance of BIOT on TUAB and LaBraM on TUEV under compression ratios ranging from $20\%$ to $80\%$. Across all compression ratios, \method consistently achieves the best performance. Even at an $80\%$ compression ratio, where competing methods degrade substantially, \method maintains stable accuracy, demonstrating strong robustness under aggressive token reduction.

\paragraph{Potential Denoising Effects of Token Pooling.}
Table~\ref{table:exp_main_6r1m} further reveals that \method frequently surpasses the uncompressed baseline (\textit{Original}). On TUAB with BIOT, it achieves an AUROC of $0.8812$ at an $80\%$ compression ratio, exceeding the original model ($0.8782$). On TUEV with LaBraM, it improves Kappa by up to $2.21\%$ at a $20\%$ compression ratio. These observations suggest that token pooling may improve representation quality by suppressing redundant or task-irrelevant EEG signals.

\begin{table}[t]
    \setlength{\tabcolsep}{2pt}
    \centering
    \begin{tabular}{lccccc}
\toprule
\multirow{2}{*}{\textbf{Variants}} & \multicolumn{2}{c}{\textbf{LaBraM / TUAB}} & & \multicolumn{2}{c}{\textbf{BIOT / TUEV}} \\
\cmidrule{2-3} \cmidrule{5-6}
 & \small AUROC & $\Delta$ & & \small Kappa & $\Delta$ \\
\midrule
\textbf{Original} & \textbf{0.8903} & — & & \textbf{0.5103} & — \\
w/ RND Group & 0.8883 & (-0.23\%) & & 0.5073 & (-0.60\%) \\
w/ RND Match & 0.8790 & (-1.27\%) & & 0.4927 & (-3.47\%) \\
w/ Simple Merge & 0.8895 & (-0.09\%) & & 0.5043 & (-1.19\%) \\
\bottomrule
\end{tabular}

    \caption{Ablation study of different components on TUAB and TUEV datasets. Performance drops compared to the original model are shown in parentheses.}
    \label{tab:ablation}
\end{table}

\subsection{Ablation Analysis}

\paragraph{Effectiveness of Components.}
Table~\ref{tab:ablation} presents the ablation study of the proposed components using LaBraM on TUAB and BIOT on TUEV. We individually replace the proposed grouping, matching, and norm-preserving merging modules with random grouping (\textit{w/ RND Group}), random matching (\textit{w/ RND Match}), and simple averaging (\textit{w/ Simple Merge}), respectively. The complete \method consistently achieves the best performance across both datasets. Here, replacing the proposed matching strategy results in the largest performance drop ($-1.27\%$ AUROC on TUAB and $-3.47\%$ Kappa on TUEV), indicating that accurate token matching is the most critical factor for effective token pooling.

\begin{figure}[t]
    \centering
    \begin{subfigure}{2.14 in}
        \includegraphics[scale=1]{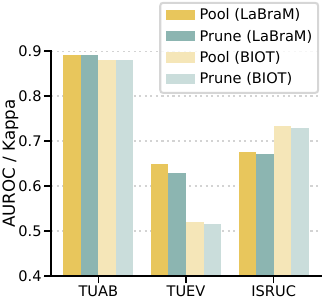}
        \subcaption{Pooling or Pruning}
        \label{fig:abl_pool_or_prune}
    \end{subfigure}
    \begin{subfigure}{1.14 in}
        \includegraphics[scale=1]{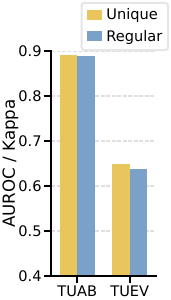}        
        \subcaption{Pivots Assignment}
        \label{fig:abl_pivots_assignment}
    \end{subfigure}
    \caption{{Ablation studies.} \textbf{(\subref{fig:abl_pool_or_prune})} Token pooling consistently outperforms token pruning across all evaluated settings. \textbf{(\subref{fig:abl_pivots_assignment})} Assigning pivots to the unique group consistently achieves better performance than treating them as regular tokens. AUROC is reported for TUAB, while Cohen's Kappa is reported for TUEV and ISRUC.}
    \label{fig:abl_combine}
\end{figure}
\begin{figure}[t]
    \centering
    \includegraphics[scale=1]{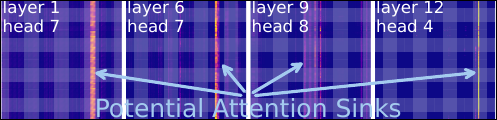}
    \caption{{Attention heatmap of LaBraM for TUEV.} The vertical lines shown in Attention heatmap implies the existence of Attention Sink. }
    \label{fig:abl_heat_map}
\end{figure}


\paragraph{Is Pooling Better Than Pruning?}
Figure~\ref{fig:abl_pool_or_prune} compares token pooling with token pruning by directly discarding redundant tokens after two-staged partitioning while keeping all other components unchanged. On TUAB, the performance gap is marginal, remaining within $0.10\%$ for both LaBraM and BIOT. However, on more challenging datasets (TUEV and ISRUC), pooling consistently outperforms pruning by $0.80\%$--$0.88\%$, with the largest improvement of $3.04\%$ Cohen's Kappa on TUEV using LaBraM. These results demonstrate that preserving redundant information through token pooling is more effective than directly pruning tokens.


\paragraph{Is It Safe to Treat Pivots as Unique Tokens?}
Figure~\ref{fig:abl_pivots_assignment} compares our default design, where pivots are assigned to the unique group, with a variant that treats pivots as regular tokens during grouping. Preserving pivots yields better performance, improving AUROC by $0.09\%$ on TUAB and Cohen's Kappa by $1.47\%$ on TUEV. A possible explanation is that high-norm pivots behave similarly to \emph{attention sinks}, serving as global structural anchors within the Transformer. Merging these tokens with others may dilute their representations and disrupt the attention pattern (Figure~\ref{fig:abl_heat_map}), whereas preserving them helps maintain model stability.
\subsection{Deployment Case Study}

\begin{figure}[t]
    \centering
    \includegraphics[scale=1]{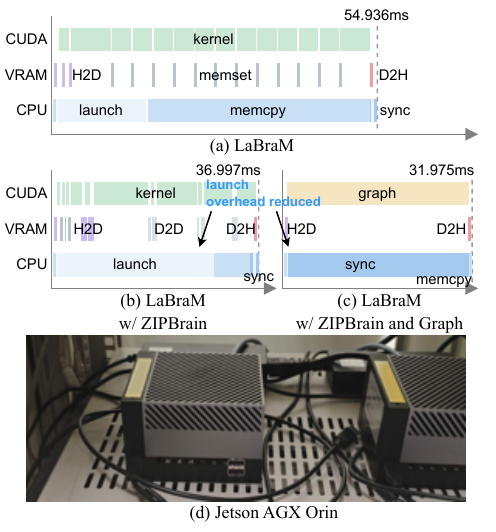}
    \caption{{Profiling timeline of LaBraM on Jetson AGX Orin via ONNX Runtime (CUDA).} {(a)} Baseline LaBraM takes 54.936\,ms. {(b)} Incorporating our proposed ZIPBrain reduces latency to 36.997\,ms. \textbf{(c)} Combining ZIPBrain with CUDA Graph further drops the total execution time to 31.975\,ms. \textbf{(d)} The Jetson AGX Orin used in the deployment experiment}    \label{fig:deploy_time_consumption}
\end{figure}

While accuracy is the gold standard, the practical utility of token compression methods ultimately depends on deployment efficiency. In this case study, we evaluate the runtime performance of \method and its compatibility with modern inference acceleration techniques. Figure~\ref{fig:deploy_time_consumption} compares the runtime profiles under standard eager execution (b) and CUDA Graph optimization (c).
Even without graph optimization, \method already exhibits high execution efficiency. Owing to its static-shape computation graph, it can be seamlessly accelerated using \textit{CUDA Graphs}, eliminating nearly all kernel launch and synchronization overheads. The optimized implementation achieves a consolidated execution time of 17.18 ms, demonstrating the suitability of our method for efficient edge deployment.



\section{Conclusion}
    \label{sec:conclusion}
In this work, we presented ZIPBrain, a training-free, plug-and-play token pooling framework that enables efficient inference for EEG foundation models by exploiting the intrinsic redundancy of low-SNR EEG signals. Extensive experiments on four representative EEG foundation models across five benchmark datasets demonstrate that ZIPBrain consistently preserves, and often improves, downstream performance while substantially reducing computational cost. These results suggest that token compression is a practical solution for deploying large-scale EEG foundation models in resource-constrained and real-time clinical settings. Future work will explore combining ZIPBrain with complementary compression techniques, such as weight pruning and quantization, and extending token compression to generative EEG foundation models.

\bibliography{BIB/References}

@article{obeid_temple_2016,
  title = {The Temple University Hospital {{EEG}} Data Corpus},
  author = {Obeid, Iyad and Picone, Joseph},
  year = 2016,
  journal = {Frontiers in Neuroscience},
  volume = {Volume 10 - 2016},
  issn = {1662-453X},
  doi = {10.3389/fnins.2016.00196}
}

@misc{dremov_computeoptimal_2025,
  title = {Compute-{{Optimal Quantization-Aware Training}}},
  author = {Dremov, Aleksandr and Grangier, David and Katharopoulos, Angelos and Hannun, Awni},
  year = 2025,
  month = sep,
  number = {arXiv:2509.22935},
  eprint = {2509.22935},
  primaryclass = {cs},
  publisher = {arXiv},
  doi = {10.48550/arXiv.2509.22935},
  urldate = {2026-02-03},
  archiveprefix = {arXiv}
}

@misc{xiao_smoothquant_2024,
  title = {{{SmoothQuant}}: {{Accurate}} and {{Efficient Post-Training Quantization}} for {{Large Language Models}}},
  shorttitle = {{{SmoothQuant}}},
  author = {Xiao, Guangxuan and Lin, Ji and Seznec, Mickael and Wu, Hao and Demouth, Julien and Han, Song},
  year = 2024,
  month = mar,
  number = {arXiv:2211.10438},
  eprint = {2211.10438},
  primaryclass = {cs},
  publisher = {arXiv},
  doi = {10.48550/arXiv.2211.10438},
  urldate = {2026-02-03},
  archiveprefix = {arXiv},
  langid = {american}
}

@inproceedings{sun_simple_2024,
  title = {A Simple and Effective Pruning Approach for Large Language Models},
  booktitle = {International Conference on Learning Representations},
  author = {Sun, Mingjie and Liu, Zhuang and Bair, Anna and Kolter, Zico},
  editor = {Kim, B. and Yue, Y. and Chaudhuri, S. and Fragkiadaki, K. and Khan, M. and Sun, Y.},
  year = 2024,
  volume = {2024},
  pages = {4942--4964}
}

@inproceedings{jiao-etal-2020-tinybert,
    title = "{T}iny{BERT}: Distilling {BERT} for Natural Language Understanding",
    author = "Jiao, Xiaoqi  and
      Yin, Yichun  and
      Shang, Lifeng  and
      Jiang, Xin  and
      Chen, Xiao  and
      Li, Linlin  and
      Wang, Fang  and
      Liu, Qun",
    editor = "Cohn, Trevor  and
      He, Yulan  and
      Liu, Yang",
    booktitle = "Findings of the Association for Computational Linguistics: EMNLP 2020",
    month = nov,
    year = "2020",
    address = "Online",
    publisher = "Association for Computational Linguistics",
    url = "https://aclanthology.org/2020.findings-emnlp.372/",
    doi = "10.18653/v1/2020.findings-emnlp.372",
    pages = "4163--4174"
}

@inproceedings{vaswani_attention_2017,
  title = {Attention Is {{All}} You {{Need}}},
  booktitle = {Advances in {{Neural Information Processing Systems}}},
  author = {Vaswani, Ashish and Shazeer, Noam and Parmar, Niki and Uszkoreit, Jakob and Jones, Llion and Gomez, Aidan N and ukasz Kaiser, {\L} and Polosukhin, Illia},
  year = 2017,
  volume = {30},
  publisher = {Curran Associates, Inc.},
  urldate = {2026-07-29}
}

@inproceedings{devlin_bert_2019,
  title = {{{BERT}}: {{Pre-training}} of {{Deep Bidirectional Transformers}} for {{Language Understanding}}},
  shorttitle = {{{BERT}}},
  booktitle = {Proceedings of the 2019 {{Conference}} of the {{North American Chapter}} of the {{Association}} for {{Computational Linguistics}}: {{Human Language Technologies}}, {{Volume}} 1 ({{Long}} and {{Short Papers}})},
  author = {Devlin, Jacob and Chang, Ming-Wei and Lee, Kenton and Toutanova, Kristina},
  editor = {Burstein, Jill and Doran, Christy and Solorio, Thamar},
  year = 2019,
  month = jun,
  pages = {4171--4186},
  publisher = {Association for Computational Linguistics},
  address = {Minneapolis, Minnesota},
  doi = {10.18653/v1/N19-1423},
  urldate = {2026-07-29}
}

@article{radford2019language,
  title={Language models are unsupervised multitask learners},
  author={Radford, Alec and Wu, Jeffrey and Child, Rewon and Luan, David and Amodei, Dario and Sutskever, Ilya and others},
  journal={OpenAI blog},
  volume={1},
  number={8},
  pages={9},
  year={2019}
}

@misc{openai2024gpt4technicalreport,
      title={GPT-4 Technical Report}, 
      author={OpenAI},
      year={2024},
      eprint={2303.08774},
      archivePrefix={arXiv},
      primaryClass={cs.CL},
      url={https://arxiv.org/abs/2303.08774}, 
}

@article{bjarkemikkelsen_eareeg_2025,
  title = {Ear-{{EEG}} Sleep Monitoring Data Sets},
  author = {Bjarke Mikkelsen, Kaare and Rezai Tabar, Yousef and R{\ae}vsb{\ae}k Birch, Laura and Lind Kappel, Simon and Bech Christensen, Christian and Dalskov Mosgaard, Lars and Otto, Marit and Christian Hemmsen, Martin and Lind Rank, Mike and Kidmose, Preben},
  year = 2025,
  month = feb,
  journal = {Scientific Data},
  volume = {12},
  number = {1},
  pages = {301},
  issn = {2052-4463},
  doi = {10.1038/s41597-025-04579-8}
}

@article{khalighi_isrucsleep_2016,
  title = {{{ISRUC-Sleep}}: {{A}} Comprehensive Public Dataset for Sleep Researchers},
  author = {Khalighi, Sirvan and Sousa, Teresa and Santos, Jos{\'e} Moutinho and Nunes, Urbano},
  year = 2016,
  journal = {Computer Methods and Programs in Biomedicine},
  volume = {124},
  pages = {180--192},
  issn = {0169-2607},
  doi = {10.1016/j.cmpb.2015.10.013}
}

@article{zyma_electroencephalograms_2019,
  title = {Electroencephalograms during Mental Arithmetic Task Performance},
  author = {Zyma, Igor and Tukaev, Sergii and Seleznov, Ivan and Kiyono, Ken and Popov, Anton and Chernykh, Mariia and Shpenkov, Oleksii},
  year = 2019,
  journal = {Data},
  volume = {4},
  number = {14},
  issn = {2306-5729},
  doi = {10.3390/data4010014}
}

@inproceedings{brown_language_2020,
  title = {Language {{Models}} Are {{Few-Shot Learners}}},
  booktitle = {Advances in {{Neural Information Processing Systems}}},
  author = {Brown, Tom and Mann, Benjamin and Ryder, Nick and Subbiah, Melanie and Kaplan, Jared D and Dhariwal, Prafulla and Neelakantan, Arvind and Shyam, Pranav and Sastry, Girish and Askell, Amanda and Agarwal, Sandhini and {Herbert-Voss}, Ariel and Krueger, Gretchen and Henighan, Tom and Child, Rewon and Ramesh, Aditya and Ziegler, Daniel and Wu, Jeffrey and Winter, Clemens and Hesse, Chris and Chen, Mark and Sigler, Eric and Litwin, Mateusz and Gray, Scott and Chess, Benjamin and Clark, Jack and Berner, Christopher and McCandlish, Sam and Radford, Alec and Sutskever, Ilya and Amodei, Dario},
  year = 2020,
  volume = {33},
  pages = {1877--1901},
  publisher = {Curran Associates, Inc.},
  urldate = {2026-07-29}
}

@misc{team_gemini_2025,
  title = {Gemini: {{A Family}} of {{Highly Capable Multimodal Models}}},
  shorttitle = {Gemini},
  author = {{Gemini Team}},
  year = 2025,
  month = may,
  number = {arXiv:2312.11805},
  eprint = {2312.11805},
  primaryclass = {cs.CL},
  publisher = {arXiv},
  doi = {10.48550/arXiv.2312.11805},
  urldate = {2026-07-29},
  archiveprefix = {arXiv}
}

@inproceedings{ankireddy_timesqueeze_2025,
  title = {{{TimeSqueeze}}: {{Dynamic Patching}} for {{Efficient Time Series Forecasting}}},
  shorttitle = {{{TimeSqueeze}}},
  booktitle = {Recent {{Advances}} in {{Time Series Foundation Models Have We Reached}} the '{{BERT Moment}}'?},
  author = {Ankireddy, Sravan Kumar and Seleznev, Nikita and Nguyen, Nam H. and Wu, Yulun and Kumar, Senthil and Huang, Furong and Bruss, C. Bayan},
  year = 2025,
  month = oct,
  urldate = {2025-12-21},
  langid = {english}
}

@inproceedings{
yang2026areiclr,
title={Are {EEG} Foundation Models Worth It? Comparative Evaluation with Traditional Decoders in Diverse {BCI} Tasks},
author={Liuyin Yang and Qiang Sun and Ang Li and Marc M. Van Hulle},
booktitle={The Fourteenth International Conference on Learning Representations},
year={2026}
}

@inproceedings{arif_hired_2025,
  title = {{{HiRED}}: {{Attention-Guided Token Dropping}} for {{Efficient Inference}} of {{High-Resolution Vision-Language Models}}},
  shorttitle = {{{HiRED}}},
  booktitle = {Proceedings of the {{AAAI Conference}} on {{Artificial Intelligence}}},
  author = {Arif, Kazi Hasan Ibn and Yoon, JinYi and Nikolopoulos, Dimitrios S. and Vandierendonck, Hans and John, Deepu and Ji, Bo},
  year = 2025,
  month = apr,
  volume = {39},
  pages = {1773--1781},
  issn = {2374-3468},
  doi = {10.1609/aaai.v39i2.32171},
  urldate = {2026-07-25},
  copyright = {Copyright (c) 2025 Association for the Advancement of Artificial Intelligence},
  langid = {english}
}

@inproceedings{bolya_token_2023,
  title = {Token {{Merging}}: {{Your ViT But Faster}}},
  shorttitle = {Token {{Merging}}},
  booktitle = {International {{Conference}} on {{Learning Representations}}},
  author = {Bolya, Daniel and Fu, Cheng-Yang and Dai, Xiaoliang and Zhang, Peizhao and Feichtenhofer, Christoph and Hoffman, Judy},
  year = 2023,
  month = mar,
  eprint = {2210.09461},
  primaryclass = {cs},
  publisher = {arXiv},
  doi = {10.48550/arXiv.2210.09461},
  urldate = {2025-12-30},
  archiveprefix = {arXiv},
  langid = {american}
}

@inproceedings{chen_image_2025,
  title = {An {{Image}} Is {{Worth}} 1/2 {{Tokens After Layer}} 2: {{Plug-and-Play Inference Acceleration}} for {{Large Vision-Language Models}}},
  shorttitle = {An {{Image}} Is {{Worth}} 1/2 {{Tokens After Layer}} 2},
  booktitle = {Computer {{Vision}} -- {{ECCV}} 2024},
  author = {Chen, Liang and Zhao, Haozhe and Liu, Tianyu and Bai, Shuai and Lin, Junyang and Zhou, Chang and Chang, Baobao},
  editor = {Leonardis, Ale{\v s} and Ricci, Elisa and Roth, Stefan and Russakovsky, Olga and Sattler, Torsten and Varol, G{\"u}l},
  year = 2025,
  volume = {15139},
  pages = {19--35},
  publisher = {Springer Nature Switzerland},
  address = {Cham},
  doi = {10.1007/978-3-031-73004-7_2},
  urldate = {2026-07-25},
  isbn = {978-3-031-73003-0 978-3-031-73004-7},
  langid = {english}
}

@inproceedings{dao_flashattention_2022,
  title = {{{FlashAttention}}: {{Fast}} and {{Memory-Efficient Exact Attention}} with {{IO-Awareness}}},
  shorttitle = {{{FlashAttention}}},
  booktitle = {Advances in {{Neural Information Processing Systems}}},
  author = {Dao, Tri and Fu, Dan and Ermon, Stefano and Rudra, Atri and R{\'e}, Christopher},
  year = 2022,
  month = dec,
  volume = {35},
  pages = {16344--16359},
  doi = {10.52202/068431-1189},
  urldate = {2026-07-25},
  langid = {english}
}

@inproceedings{dao_flashattention2_2024,
  title = {{{FlashAttention-2}}: {{Faster Attention}} with {{Better Parallelism}} and {{Work Partitioning}}},
  shorttitle = {{{FlashAttention-2}}},
  booktitle = {International {{Conference}} on {{Learning Representations}}},
  author = {Dao, Tri},
  year = 2024,
  month = may,
  volume = {2024},
  pages = {35549--35562},
  urldate = {2026-07-25},
  langid = {english}
}

@inproceedings{dhouib_pact_2025,
  title = {{{PACT}}: {{Pruning}} and {{Clustering-Based Token Reduction}} for {{Faster Visual Language Models}}},
  shorttitle = {{{PACT}}},
  booktitle = {Proceedings of the {{IEEE}}/{{CVF Conference}} on {{Computer Vision}} and {{Pattern Recognition}}},
  author = {Dhouib, Mohamed and Buscaldi, Davide and Vanier, Sonia and Shabou, Aymen},
  year = 2025,
  pages = {14582--14592},
  urldate = {2026-07-25},
  langid = {english}
}

@inproceedings{fang_prune_2025,
  title = {Prune {{Redundancy}}, {{Preserve Essence}}: {{Vision Token Compression}} in {{VLMs}} via {{Synergistic Importance-Diversity}}},
  shorttitle = {Prune {{Redundancy}}, {{Preserve Essence}}},
  booktitle = {The {{Fourteenth International Conference}} on {{Learning Representations}}},
  author = {Fang, Zhengyao and Lyu, Pengyuan and Zhang, Chengquan and Lu, Guangming and Yu, Jun and Pei, Wenjie},
  year = 2025,
  month = oct,
  urldate = {2026-07-25},
  langid = {english}
}

@inproceedings{jiang_large_2024,
  title = {Large {{Brain Model}} for {{Learning Generic Representations}} with {{Tremendous EEG Data}} in {{BCI}}},
  booktitle = {International {{Conference}} on {{Learning Representations}}},
  author = {Jiang, Wei-Bang and Zhao, Liming and Lu, Bao-liang},
  editor = {Kim, B. and Yue, Y. and Chaudhuri, S. and Fragkiadaki, K. and Khan, M. and Sun, Y.},
  year = 2024,
  volume = {2024},
  pages = {16405--16426}
}

@inproceedings{kim_token_2024,
  title = {Token {{Fusion}}: {{Bridging}} the {{Gap}} between {{Token Pruning}} and {{Token Merging}}},
  shorttitle = {Token {{Fusion}}},
  booktitle = {2024 {{IEEE}}/{{CVF Winter Conference}} on {{Applications}} of {{Computer Vision}} ({{WACV}})},
  author = {Kim, Minchul and Gao, Shangqian and Hsu, Yen-Chang and Shen, Yilin and Jin, Hongxia},
  year = 2024,
  month = jan,
  pages = {1372--1381},
  publisher = {IEEE},
  address = {Waikoloa, HI, USA},
  doi = {10.1109/WACV57701.2024.00141},
  urldate = {2026-01-07},
  copyright = {https://doi.org/10.15223/policy-029},
  isbn = {979-8-3503-1892-0},
  langid = {english}
}

@inproceedings{liang_evit_2021,
  title = {{{EViT}}: {{Expediting Vision Transformers}} via {{Token Reorganizations}}},
  shorttitle = {{{EViT}}},
  booktitle = {International {{Conference}} on {{Learning Representations}}},
  author = {Liang, Youwei and Ge, Chongjian and Tong, Zhan and Song, Yibing and Wang, Jue and Xie, Pengtao},
  year = 2021,
  month = oct,
  urldate = {2026-07-25},
  langid = {english}
}

@inproceedings{ma_codebrain_2025,
  title = {{{CodeBrain}}: {{Bridging Decoupled Tokenizer}} and {{Multi-Scale Architecture}} for {{EEG Foundation Model}}},
  shorttitle = {{{CodeBrain}}},
  booktitle = {The {{Fourteenth International Conference}} on {{Learning Representations}}},
  author = {Ma, Jingying and Wu, Feng and Lin, Qika and Xing, Yucheng and Liu, Chenyu and Jia, Ziyu and Feng, Mengling},
  year = 2025,
  month = oct,
  urldate = {2026-07-14},
  langid = {english}
}

@inproceedings{pradeepkumar_tokenizing_2025,
  title = {Tokenizing {{Single-Channel EEG}} with {{Time-Frequency Motif Learning}}},
  booktitle = {The {{Fourteenth International Conference}} on {{Learning Representations}}},
  author = {Pradeepkumar, Jathurshan and Piao, Xihao and Chen, Zheng and Sun, Jimeng},
  year = 2025,
  month = oct,
  urldate = {2026-07-14},
  langid = {english}
}

@inproceedings{rao_dynamicvit_2021a,
  title = {{{DynamicViT}}: {{Efficient Vision Transformers}} with {{Dynamic Token Sparsification}}},
  shorttitle = {{{DynamicViT}}},
  booktitle = {Advances in {{Neural Information Processing Systems}}},
  author = {Rao, Yongming and Zhao, Wenliang and Liu, Benlin and Lu, Jiwen and Zhou, Jie and Hsieh, Cho-Jui},
  year = 2021,
  volume = {34},
  pages = {13937--13949},
  publisher = {Curran Associates, Inc.},
  urldate = {2026-07-25}
}

@inproceedings{AAAI2025SODOR,
  author={Chen, Zheng and Matsubara, Yasuko and Sakurai, Yasushi and Sun, Jimeng},
  title={Long-Term {EEG} Partitioning for Seizure Onset Detection},
  booktitle={Proceedings of the AAAI Conference on Artificial Intelligence},
  year={2025},
  pages={14221--14229}
}

@inproceedings{tegon_femba_2025,
  title = {{{FEMBA}}: {{Efficient}} and {{Scalable EEG Analysis}} with a {{Bidirectional Mamba Foundation Model}}},
  shorttitle = {{{FEMBA}}},
  booktitle = {2025 47th {{Annual International Conference}} of the {{IEEE Engineering}} in {{Medicine}} and {{Biology Society}} ({{EMBC}})},
  author = {Tegon, Anna and Ingolfsson, Thorir Mar and Wang, Xiaying and Benini, Luca and Li, Yawei},
  year = 2025,
  month = jul,
  pages = {1--7},
  issn = {2694-0604},
  doi = {10.1109/EMBC58623.2025.11252697},
  urldate = {2026-07-14}
}

@inproceedings{wang_cbramod_2024,
  title = {{{CBraMod}}: {{A Criss-Cross Brain Foundation Model}} for {{EEG Decoding}}},
  shorttitle = {{{CBraMod}}},
  booktitle = {The {{Thirteenth International Conference}} on {{Learning Representations}}},
  author = {Wang, Jiquan and Zhao, Sha and Luo, Zhiling and Zhou, Yangxuan and Jiang, Haiteng and Li, Shijian and Li, Tao and Pan, Gang},
  year = 2024,
  month = oct,
  urldate = {2026-07-14},
  langid = {english}
}

@inproceedings{wang_eegpt_2024,
  title = {{{EEGPT}}: {{Pretrained Transformer}} for {{Universal}} and {{Reliable Representation}} of {{EEG Signals}}},
  shorttitle = {{{EEGPT}}},
  booktitle = {The {{Thirty-eighth Annual Conference}} on {{Neural Information Processing Systems}}},
  author = {Wang, Guangyu and Liu, Wenchao and He, Yuhong and Xu, Cong and Ma, Lin and Li, Haifeng},
  year = 2024,
  month = nov,
  urldate = {2025-12-15},
  langid = {english}
}

@misc{wang_linformer_2020,
  title = {Linformer: {{Self-Attention}} with {{Linear Complexity}}},
  shorttitle = {Linformer},
  author = {Wang, Sinong and Li, Belinda Z. and Khabsa, Madian and Fang, Han and Ma, Hao},
  year = 2020,
  month = jun,
  number = {arXiv:2006.04768},
  eprint = {2006.04768},
  primaryclass = {cs},
  publisher = {arXiv},
  doi = {10.48550/arXiv.2006.04768},
  urldate = {2026-05-03},
  archiveprefix = {arXiv}
}

@inproceedings{wen_stop_2025,
  title = {Stop {{Looking}} for ``{{Important Tokens}}'' in {{Multimodal Language Models}}: {{Duplication Matters More}}},
  shorttitle = {Stop {{Looking}} for ``{{Important Tokens}}'' in {{Multimodal Language Models}}},
  booktitle = {Proceedings of the 2025 {{Conference}} on {{Empirical Methods}} in {{Natural Language Processing}}},
  author = {Wen, Zichen and Gao, Yifeng and Wang, Shaobo and Zhang, Junyuan and Zhang, Qintong and Li, Weijia and He, Conghui and Zhang, Linfeng},
  editor = {Christodoulopoulos, Christos and Chakraborty, Tanmoy and Rose, Carolyn and Peng, Violet},
  year = 2025,
  month = nov,
  pages = {9961--9980},
  publisher = {Association for Computational Linguistics},
  address = {Suzhou, China},
  doi = {10.18653/v1/2025.emnlp-main.505},
  urldate = {2026-02-26},
  isbn = {979-8-89176-332-6},
  langid = {american}
}

@inproceedings{yang_biot_2023,
  title = {{{BIOT}}: {{Biosignal Transformer}} for {{Cross-data Learning}} in the {{Wild}}},
  booktitle = {Advances in {{Neural Information Processing Systems}}},
  author = {Yang, Chaoqi and Westover, M and Sun, Jimeng},
  editor = {Oh, A. and Naumann, T. and Globerson, A. and Saenko, K. and Hardt, M. and Levine, S.},
  year = 2023,
  volume = {36},
  pages = {78240--78260},
  publisher = {Curran Associates, Inc.}
}

@inproceedings{ye_fit_2025,
  title = {Fit and {{Prune}}: {{Fast}} and {{Training-free Visual Token Pruning}} for {{Multi-modal Large Language Models}}},
  shorttitle = {Fit and {{Prune}}},
  booktitle = {Proceedings of the {{AAAI Conference}} on {{Artificial Intelligence}}},
  author = {Ye, Weihao and Wu, Qiong and Lin, Wenhao and Zhou, Yiyi},
  year = 2025,
  month = apr,
  volume = {39},
  pages = {22128--22136},
  issn = {2374-3468},
  doi = {10.1609/aaai.v39i21.34366},
  urldate = {2026-02-26},
  copyright = {Copyright (c) 2025 Association for the Advancement of Artificial Intelligence},
  langid = {english}
}

@inproceedings{zhang_textvisual_2025,
  title = {Beyond {{Text-Visual Attention}}: {{Exploiting Visual Cues}} for {{Effective Token Pruning}} in {{VLMs}}},
  shorttitle = {Beyond {{Text-Visual Attention}}},
  booktitle = {Proceedings of the {{IEEE}}/{{CVF International Conference}} on {{Computer Vision}}},
  author = {Zhang, Qizhe and Cheng, Aosong and Lu, Ming and Zhang, Renrui and Zhuo, Zhiyong and Cao, Jiajun and Guo, Shaobo and She, Qi and Zhang, Shanghang},
  year = 2025,
  pages = {20857--20867},
  urldate = {2026-07-25},
  langid = {english}
}

@article{zhou_adaptive_2026,
  title = {Adaptive {{Segmentation}} of {{EEG}} for {{Machine Learning Applications}}},
  author = {Zhou, Johnson and West, Joseph and Ehinger, Krista A. and Ren, Zhenming and John, Sam E. and Grayden, David B.},
  year = 2026,
  journal = {IEEE Journal of Biomedical and Health Informatics},
  pages = {1--14},
  issn = {2168-2208},
  doi = {10.1109/JBHI.2026.3670481},
  urldate = {2026-07-14}
}

\section{Appendix}
\setcounter{secnumdepth}{2}
\appendix
    \label{sec:appendix}
    \section{Additional Related Work}
\noindent \textbf{Efficient Transformers.} The computational overhead of large-scale Transformer models has catalyzed several distinct lines of efficiency research. \textbf{Quantization}~\cite{dremov_computeoptimal_2025, xiao_smoothquant_2024} compresses models by lowering the numerical precision of weights and activations. \textbf{Knowledge distillation}~\cite{jiao-etal-2020-tinybert} improves inference efficiency by training compact student architectures to mimic the representations of larger teacher models. \textbf{Weight pruning}~\cite{sun_simple_2024} eliminates near-zero parameters, which—when paired with sparse storage formats and hardware zero-skipping—alleviates both compute and memory constraints. \textbf{Linear attention approximations}~\cite{wang_linformer_2020} aim to bypass the quadratic complexity bottleneck, albeit often at the cost of reduced feature dimensionality and diminished model expressiveness. Furthermore, \textbf{FlashAttention}~\cite{dao_flashattention_2022, dao_flashattention2_2024} represents a hardware-aware breakthrough that accelerates exact attention execution, serving as a fundamental component in modern Transformer architectures.

\section{Additional Experimental Details}

\subsection{Dataset Accessibility}
All five evaluation benchmarks are publicly accessible online:
\begin{itemize}
    \item TUAB and TUEV \cite{obeid_temple_2016}: \url{https://isip.piconepress.com/projects/nedc/html/tuh_eeg/}
    \item ISRUC \cite{khalighi_isrucsleep_2016}: \url{https://sleeptight.isr.uc.pt/}
    \item EarEEG \cite{bjarkemikkelsen_eareeg_2025}: \url{https://openneuro.org/datasets/ds005178}
    \item EEGMAT \cite{zyma_electroencephalograms_2019}: \url{https://physionet.org/content/eegmat/1.0.0/}
\end{itemize}

\subsection{Data Preprocessing Procedure}
The TUAB and TUEV datasets are preprocessed strictly following the scripts provided by their respective baseline implementations, whereas EarEEG adheres to the preprocessing protocol defined in TFM-Tokenizer.

For ISRUC, each epoch is formatted as an individual sample comprising six EEG channels (\texttt{F3-A2, C3-A2, O1-A2, F4-A1, C4-A1, O2-A1}), using \texttt{\{\}\_1.txt} as label files. Data partitions are constructed at the subject level: Subjects 1--80 are allocated to training, 81--90 to validation, and 91--100 to testing.

For EEGMAT, signals are resampled from 500~Hz to 200~Hz and segmented into non-overlapping 2-second patches (with the initial 2 seconds discarded). The subject-level split assigns Subject00--Subject25 to training, Subject26--Subject30 to validation, and Subject31--Subject35 to testing.

\subsection{Evaluation Metrics}
The evaluation procedures and toolchains are established as follows:
\begin{itemize}
    \item Accuracy-related classification metrics are computed using \texttt{scikit-learn}.
    \item Computational FLOPs are defined as twice the count of Multiply-Accumulate operations (MACs) and profiled via \texttt{onnx-tool} after converting model checkpoints to \texttt{onnx} format.
    \item Inference latency is measured using execution logs captured by NVIDIA Nsight Systems, evaluated with a batch size of 32 over 32 warm-up and measurement iterations.
\end{itemize}

\subsection{Hyperparameters for Fine-Tuning}
We fine-tune all four foundation models independently across all five downstream tasks and subsequently evaluate token compression techniques on top of these fine-tuned checkpoints. To ensure maximum reproducibility, we utilize the official code repositories and public checkpoints. The exact hyperparameter configurations used during fine-tuning are summarized in Table~\ref{tab:finetune_hyps}.

The source code and pre-trained checkpoints are available at the following GitHub repositories:
\begin{itemize}
    \item BIOT: \url{https://github.com/ycq091044/BIOT}
    \item LaBraM: \url{https://github.com/935963004/LaBraM}
    \item EEGPT: \url{https://github.com/BINE022/EEGPT}
    \item TFM-Tokenizer: \url{https://github.com/Jathurshan0330/TFM-Tokenizer}
\end{itemize}

\begin{table*}[ht]
    \centering
    \small
    \caption{Hyperparameter configurations for fine-tuning downstream EEG foundation models. Pre-trained checkpoints for TUAB and TUEV are provided by TFM-Tokenizer.}
    \label{tab:finetune_hyps}
    \begin{tabular}{l|ccccc|ccccc}
        \toprule
        Model       & \multicolumn{5}{c|}{LaBraM} & \multicolumn{5}{c}{EEGPT} \\
        Dataset     & TUAB & TUEV & ISRUC & EarEEG & EEGMAT & TUAB & TUEV & ISRUC & EarEEG & EEGMAT \\
        \midrule
        layers      & \multicolumn{5}{c|}{12} & \multicolumn{5}{c}{8} \\
        initial tokens 
                    & 230 & 115 & 180 & 120 & 64 & 20 & 20 & 6 & 4 & 16 \\
        \midrule
        optimizer   & \multicolumn{5}{c|}{AdamW} & \multicolumn{5}{c}{AdamW} \\
        scheduler   & \multicolumn{5}{c|}{Cosine} & \multicolumn{2}{c|}{Cosine} & \multicolumn{3}{c}{OneCycleLR} \\
        seed        & \multicolumn{5}{c|}{0} & \multicolumn{2}{c|}{0} & \multicolumn{3}{c}{7} \\
        Batch Size  & \multicolumn{4}{c|}{256} & 377 & \multicolumn{2}{c|}{256} & \multicolumn{3}{c}{32} \\
        epoch       & \multicolumn{5}{c|}{50} & \multicolumn{2}{c|}{50} & \multicolumn{3}{c}{40} \\
        warm-up     & \multicolumn{5}{c|}{1} & \multicolumn{2}{c|}{5} & \multicolumn{3}{c}{8} \\
        lr          & \multicolumn{5}{c|}{5e-4} & \multicolumn{2}{c|}{5e-4} & \multicolumn{3}{c}{4e-4}\\
        weight decay& \multicolumn{5}{c|}{0.05} & \multicolumn{2}{c|}{0.05} & \multicolumn{3}{c}{0.01} \\
        layer decay & \multicolumn{5}{c|}{0.65} & \multicolumn{2}{c|}{0.65} & \multicolumn{3}{c}{0} \\
        drop path   & \multicolumn{5}{c|}{0.1} & 0.1 & \multicolumn{1}{c|}{0.2} & \multicolumn{3}{c}{0} \\
        \midrule
        \midrule
        Model       & \multicolumn{5}{c|}{BIOT} & \multicolumn{5}{c}{TFM-Tokenizer} \\
        Dataset     & TUAB & TUEV & ISRUC & EarEEG & EEGMAT & TUAB & TUEV & ISRUC & EarEEG & EEGMAT \\
        \midrule
        layers      & \multicolumn{5}{c|}{4} & \multicolumn{5}{c}{4} \\
        initial tokens 
                    & 304 & 144 & 354 & 236 & 112 & 304 & 144 & 354 & 236 & 112 \\
        \midrule
        optimizer   & \multicolumn{5}{c|}{Adam} & \multicolumn{2}{c|}{/} & \multicolumn{3}{c}{AdamW} \\
        scheduler   & \multicolumn{5}{c|}{/} & \multicolumn{2}{c|}{/} & \multicolumn{3}{c}{Custom} \\
        seed        & \multicolumn{5}{c|}{12345} & \multicolumn{2}{c|}{/} & \multicolumn{3}{c}{5} \\
        Batch Size  & \multicolumn{4}{c|}{512} & 754 & \multicolumn{2}{c|}{/} & \multicolumn{2}{c|}{256} & 377 \\
        epoch       & \multicolumn{5}{c|}{25} & \multicolumn{2}{c|}{/} & \multicolumn{3}{c}{50} \\
        warm-up     & \multicolumn{5}{c|}{/} & \multicolumn{2}{c|}{/} & \multicolumn{3}{c}{5} \\
        lr          & \multicolumn{5}{c|}{5e-4} & \multicolumn{2}{c|}{/} & \multicolumn{3}{c}{1e-3}\\
        \bottomrule
    \end{tabular}
\end{table*}

\subsection{Baseline Implementations}
Official implementations of the three baseline token compression methods are obtained from their public GitHub repositories. Our unified evaluation framework is built upon ToMe's implementation:
\begin{itemize}
    \item ToMe: \url{https://github.com/facebookresearch/ToMe}
    \item EViT: \url{https://github.com/youweiliang/evit}
    \item DART: \url{https://github.com/ZichenWen1/DART}
\end{itemize}

\subsection{Implementation of the Proposed Method}
The complete execution pipeline of \method is formally detailed as pseudocode in Algorithm~\ref{alg:kidd_left1s}.

\subsection{Hyperparameter Optimization Procedure}
\label{supp:hyperparameter_optimization}

We define the hyperparameter search space for our proposed method as follows:
\begin{itemize}
    \item \textbf{Module Representations ($\mathbf{Z}_1, \mathbf{Z}_2, \mathbf{Z}_3$):} 
    The tensor representations used in Pivot Selection ($\mathbf{Z}_1$), Redundant Token Identification ($\mathbf{Z}_2$), and Target Token Matching ($\mathbf{Z}_3$) are selected from the candidate set $\{\mathbf{Q}, \mathbf{K}, \mathbf{V}, \hat{\mathbf{X}}\}$.

    \item \textbf{Multi-Head Aggregation Strategy:} 
    In multi-head self-attention, aggregating token representations across attention heads is non-trivial. For multi-head tensors ($\mathbf{Q}, \mathbf{K}, \mathbf{V}$), we evaluate two aggregation strategies: \textit{Pooled Representation} (averaging representations across heads prior to score computation, designated by suffix $S$) and \textit{Pooled Score} (averaging computed scores across heads, designated by suffix $H$). Together with the single-tensor $\hat{\mathbf{X}}$, each module admits 7 candidate configurations.

    \item \textbf{Pivot Proportion ($\rho$):} 
    The proportion of retained pivot tokens is governed by $\rho \in (0, 1]$. To make grid search tractable, we discretize this space into 11 candidate values: $\rho \in \{0.0, 0.1, \dots, 1.0\}$. Here, $\rho=0.0$ represents a boundary case where exactly one pivot token is selected.
\end{itemize}

The exhaustive search space encompasses $7^3 \times 11 = 3{,}773$ candidate configurations. To mitigate computational overhead, we adopt a structured two-stage grid search procedure:
\begin{enumerate}
    \item \textbf{Stage 1 (Joint Search for $\mathbf{Z}_1, \mathbf{Z}_2$, and $\rho$):} We fix the matching representation $\mathbf{Z}_3 = \hat{\mathbf{X}}$ and perform a grid search over 27 candidate configurations for $\{\mathbf{Z}_1, \mathbf{Z}_2\}$ across all 11 pivot ratios, forming an initial search space of $27 \times 11 = 297$ combinations. The 27 candidate configurations for $\{\mathbf{Z}_1, \mathbf{Z}_2\}$ are:
    \begin{itemize}
        \item $(\hat{\mathbf{X}}, \hat{\mathbf{X}}), (\hat{\mathbf{X}}, \mathbf{QS}), (\hat{\mathbf{X}}, \mathbf{KS}), (\hat{\mathbf{X}}, \mathbf{VS}), $
        \item $(\hat{\mathbf{X}}, \mathbf{QH}), (\hat{\mathbf{X}}, \mathbf{KH}), (\hat{\mathbf{X}}, \mathbf{VH}),$
        \item $(\mathbf{VS}, \hat{\mathbf{X}}), (\mathbf{VS}, \mathbf{QS}), (\mathbf{VS}, \mathbf{KS}), (\mathbf{VS}, \mathbf{VS}),$
        \item $(\mathbf{VH}, \hat{\mathbf{X}}), (\mathbf{VH}, \mathbf{QH}), (\mathbf{VH}, \mathbf{KH}), (\mathbf{VH}, \mathbf{VH}),$
        \item $(\mathbf{QS}, \hat{\mathbf{X}}), (\mathbf{QS}, \mathbf{QS}), (\mathbf{QS}, \mathbf{VS}),$
        \item $(\mathbf{QH}, \hat{\mathbf{X}}), (\mathbf{QH}, \mathbf{QH}), (\mathbf{QH}, \mathbf{VH}),$
        \item $(\mathbf{KS}, \hat{\mathbf{X}}), (\mathbf{KS}, \mathbf{KS}), (\mathbf{KS}, \mathbf{VS}),$
        \item $(\mathbf{KH}, \hat{\mathbf{X}}), (\mathbf{KH}, \mathbf{KH}), (\mathbf{KH}, \mathbf{VH}),$
    \end{itemize}
    \item \textbf{Stage 2 (Refinement for $\mathbf{Z}_3$):} We identify the top-4 performing configurations from Stage 1 and optimize $\mathbf{Z}_3$ across its remaining candidates.
\end{enumerate}
All main experimental results reported in the manuscript are based on the optimal hyperparameter configuration derived from this two-stage optimization. Because the hyperparameter optimization procedure is defined deterministically, the final results are free from random variation.

\section{Computational Complexity Analysis}

\begin{algorithm}[t]
\caption{ZIPBrain Token Compression Pipeline}
\label{alg:kidd_left1s}
\begin{algorithmic}[1]
\renewcommand{\algorithmicrequire}{\textbf{Input:}}
\renewcommand{\algorithmicensure}{\textbf{Output:}}

\REQUIRE Token features $\hat{\mathbf{X}} \in \mathbb{R}^{N \times d}$; Metric representations $\mathbf{Z}_1, \mathbf{Z}_2, \mathbf{Z}_3 \in \mathbb{R}^{N \times H \times d_h}$; Token budget $r$; Pivot proportion $\rho$.
\ENSURE Merged token features $\hat{\mathbf{X}}' \in \mathbb{R}^{(N-r) \times d}$.

\STATE $N_{\text{pivot}} \leftarrow \min\left(\max\left(\lceil (N - r) \cdot \rho \rceil, 1\right), N - r\right)$ 
\STATE \textbf{// Step 1: Pivot Token Selection}
\STATE $S_0 \leftarrow \text{Mean}_{\text{heads}}(\|\mathbf{Z}_1\|_2)$
\STATE $I_{\text{pivot}} \leftarrow \text{TopK\_Indices}(S_0, N_{\text{pivot}})$
\STATE $I_{\text{non\_pivot}} \leftarrow \{1, \dots, N\} \setminus I_{\text{pivot}}$

\STATE \textbf{// Step 2: Redundancy Scoring \& Source Selection}
\STATE $\hat{\mathbf{Z}}_2 \leftarrow \text{Normalize}(\mathbf{Z}_2, L_2)$
\STATE $\mathbf{T}_{\text{pivot}} \leftarrow \text{Gather}(\hat{\mathbf{Z}}_2, I_{\text{pivot}})$, \quad $\mathbf{T}_{\text{non\_pivot}} \leftarrow \text{Gather}(\hat{\mathbf{Z}}_2, I_{\text{non\_pivot}})$
\STATE $S_{\text{dup}} \leftarrow \text{Mean}_{\text{heads}}(\mathbf{T}_{\text{non\_pivot}} \cdot \text{Mean}_{\text{tokens}}( \mathbf{T}_{\text{pivot}}) )$
\STATE $I_{\text{src\_rel}} \leftarrow \text{TopK\_Indices}(S_{\text{dup}}, r)$
\STATE $I_{\text{src}} \leftarrow \text{Gather}(I_{\text{non\_pivot}}, I_{\text{src\_rel}})$ 
\STATE $I_{\text{left}} \leftarrow I_{\text{pivot}} \cup (I_{\text{non\_pivot}} \setminus I_{\text{src}})$ 

\STATE \textbf{// Step 3: Target Token Matching}
\STATE $\hat{\mathbf{Z}}_3 \leftarrow \text{Normalize}(\mathbf{Z}_3, L_2)$
\STATE $\mathbf{T}_{\text{src}} \leftarrow \text{Gather}(\hat{\mathbf{Z}}_3, I_{\text{src}})$, \quad $\mathbf{T}_{\text{left}} \leftarrow \text{Gather}(\hat{\mathbf{Z}}_3, I_{\text{left}})$
\STATE $S_{\text{tgt}} \leftarrow \text{Mean}_{\text{heads}}(\mathbf{T}_{\text{left}} \cdot \mathbf{T}_{\text{src}}^T)$
\STATE $I_{\text{sim}} \leftarrow \argmax(S_{\text{tgt}}, \text{dim}=\text{left})$
\STATE $I_{\text{tar}} \leftarrow \text{Gather}(I_{\text{left}}, I_{\text{sim}})$ 

\STATE \textbf{// Step 4: Feature Aggregation \& Rescaling}
\STATE \textbf{Define} $\text{Merge}(\mathbf{V}, \text{op})$:
\STATE \quad $\mathbf{V}_{\text{tar}} \leftarrow \text{ScatterReduce}(\mathbf{V}, \text{indices}=I_{\text{tar}}, \text{src}=\mathbf{V}[I_{\text{src}}], \text{reduce}=\text{op})$
\STATE \quad \textbf{return} $\mathbf{V}_{\text{tar}}[I_{\text{left}}]$

\STATE $\mathbf{L} \leftarrow \|\hat{\mathbf{X}}\|_2$
\STATE $\mathbf{L}' \leftarrow \text{Merge}(\mathbf{L}, \text{``amax''})$ 
\STATE $\hat{\mathbf{X}}_{\text{merged}} \leftarrow \text{Merge}(\hat{\mathbf{X}}, \text{``sum''})$
\STATE $\hat{\mathbf{X}}' \leftarrow \frac{\hat{\mathbf{X}}_{\text{merged}}}{\|\hat{\mathbf{X}}_{\text{merged}}\|_2} \odot \mathbf{L}'$ 

\RETURN $\hat{\mathbf{X}}'$
\end{algorithmic}
\end{algorithm}

Let $N$ denote the number of input tokens, $d$ the feature dimension, $H$ the number of attention heads, and $d_h=d/H$ the feature dimension per head. Let $r$ denote the number of removed tokens and $\rho$ the pivot proportion, yielding
\[
N_{\mathrm{pivot}}=\left\lceil (N-r)\rho \right\rceil.
\]

Algorithm~\ref{alg:kidd_left1s} consists of four primary stages:

\paragraph{Pivot Selection.}
The pivot score is calculated by computing the $\ell_2$-norm of each token representation and averaging across attention heads, incurring a complexity of
\[
\mathcal{O}(Nd).
\]
Selecting the top-$N_{\mathrm{pivot}}$ pivot indices introduces negligible computational cost compared to feature transformation.

\paragraph{Redundant Token Identification.}
Rather than computing pairwise similarities between every pivot and non-pivot token, \method aggregates all pivot representations into a single prototype via mean pooling. Non-pivot tokens are subsequently compared against this single prototype, reducing complexity to linear scale:
\[
\mathcal{O}(Nd).
\]

\paragraph{Target Token Matching.}
The matching stage computes pairwise similarity scores between the $r$ selected source tokens and the $(N-r)$ retained tokens, resulting in
\[
\mathcal{O}(r(N-r)d).
\]

\paragraph{Feature Aggregation.}
The scatter-reduce operation and subsequent feature norm rescaling are both linear in the token count:
\[
\mathcal{O}(Nd).
\]

\paragraph{Overall Complexity.}
Combining all stages yields an overall computational complexity of
\[
\mathcal{O}\!\left(Nd + r(N-r)d\right),
\]
which is bounded by $\mathcal{O}(N^2 d)$ in the worst-case scenario.

Because redundancy scoring and feature aggregation operate with linear complexity relative to the sequence length, the quadratic term in \method originates solely from the target matching stage.

\section{Additional Results}
In addition to the AUROC, PR-AUC, Cohen's Kappa, and Weighted-F1 metrics reported in the main paper, we provide Accuracy (Acc) and Balanced Accuracy (BAcc) here for completeness. Table~\ref{tab:4_model_1_ratio} details the classification performance under the maximum available compression ratios across models. Table~\ref{tab:2_model_4_ratio} reports performance across varying compression ratios for TUAB evaluated on BIOT and TUEV evaluated on LaBraM.

\begin{table*}[ht]
    \centering
    \begin{threeparttable}
\newcolumntype{H}{>{\setbox0=\hbox\bgroup}c<{\egroup}@{}}
\begin{tabular}{ll Hcc Hcc | Hcc Hcc Hcc}
    \toprule
    \multirow{2}{*}{Model} & \multirow{2}{*}{Methods} & \multicolumn{3}{c}{TUAB} & \multicolumn{3}{c}{EEGMAT} & \multicolumn{3}{c}{TUEV} & \multicolumn{3}{c}{ISRUC} & \multicolumn{3}{c}{EarEEG} \\
    \cmidrule(lr){3-5} \cmidrule(lr){6-8} \cmidrule(lr){9-11} \cmidrule(lr){12-14} \cmidrule(lr){15-17}
    & & Hyper-Params &  Acc &  BAcc & Hyper-Params &  Acc &  BAcc & Hyper-Params & Acc & BAcc & Hyper-Params & Acc & BAcc & Hyper-Params & Acc & BAcc \\

    \midrule
    \multirow{6.5}{*}{ LaBraM} 
    & Original\footnotemark[1] &  & 0.8221 & 0.8206 &  & 0.5655 & 0.5655 &  & 0.8238 & 0.6585 &  & 0.8003 & 0.7717 &  & 0.5951 & 0.4463 \\
    \cmidrule(lr){2-17}
    & EVIT &  & 0.7716 & 0.7687 &  & 0.5793 & 0.5793 &  & 0.8120 & 0.4357 &  & 0.6908 & 0.6641 &  & 0.4226 & 0.3566 \\
    & ToMe &  & \underline{0.8108} & 0.8036 &  & 0.5724 & 0.5724 &  & 0.8113 & 0.5975 &  & 0.6982 & 0.6684 &  & 0.4728 & 0.3869 \\
    & ToFU &  & \textbf{0.8150} & \textbf{0.8085} &  & 0.5828 & 0.5828 &  & 0.8133 & 0.6025 &  & 0.7183 & 0.6866 &  & 0.4759 & 0.3855 \\
    & DART &  & 0.7918 & 0.7918 &  & 0.5621 & 0.5621 &  & 0.7642 & 0.5468 &  & 0.7445 & 0.7020 &  & 0.5326 & 0.3934 \\
    & Ours & x,v,4,x,10 & 0.8041 & \underline{0.8053} & x,q,3,x,10 & \textbf{0.6172} & \textbf{0.6172} & k,k,9,k,10 & \textbf{0.8269} & \textbf{0.6212} & q,v,1,x,10 & \textbf{0.7518} & \textbf{0.7244} & x,vh,1,kh,10 & \textbf{0.5802} & \textbf{0.4607} \\
    
    \midrule
    \multirow{6.5}{*}{EEGPT} 
    & Original\footnotemark[1] &  & 0.8012 & 0.7992 &  & 0.6034 & 0.6034 &  & 0.8014 & 0.5411 &  & 0.7383 & 0.7141 &  & 0.6196 & 0.4675 \\
    \cmidrule(lr){2-17}
    & EVIT\textsuperscript{mean} &  & \textbf{0.8000} & 0.7982 &  & 0.6138 & 0.6138 &  & 0.8053 & 0.5316 &  & 0.7347 & 0.7112 &  & 0.6009 & 0.4513 \\
    & ToMe &  & 0.7998 & 0.7980 &  & \textbf{0.6207} & \textbf{0.6207} &  & 0.8008 & 0.5315 &  & 0.7245 & 0.7028 &  & 0.6056 & 0.4457 \\
    & ToFU &  & 0.7995 & 0.7977 &  & \textbf{0.6207} & \textbf{0.6207} &  & 0.8011 & 0.5311 &  & 0.7269 & 0.7041 &  & 0.6050 & 0.4464 \\
    & DART &  & \underline{0.7999} & \textbf{0.7985} &  & \underline{0.6172} & \underline{0.6172} &  & \textbf{0.8029} & 0.5378 &  & \textbf{0.7424} & \textbf{0.7162} &  & \textbf{0.6131} & 0.4530 \\
    & Ours & kh,x,0,x,10 & 0.7989 & 0.7974 & x,v,0,v,10 & 0.6138 & 0.6138 & q,q,10,q,10 & \underline{0.8018} & \textbf{0.5527} & x,q,5,qh,10 & \underline{0.7402} & \underline{0.7143} & vh,qh,6,v,10 & \underline{0.6121} & \textbf{0.4571} \\

    \midrule
    \multirow{6.5}{*}{BIOT} 
    & Original\footnotemark[1] &  & 0.8081 & 0.8009 &  & 0.6276 & 0.6276 &  & 0.7274 & 0.5059 &  & 0.8132 & 0.7776 &  & 0.5802 & 0.4136 \\
    \cmidrule(lr){2-17}
    & EVIT\textsuperscript{mean} &  & 0.7913 & 0.7900 &  & 0.6138 & 0.6138 &  & 0.7007 & 0.4959 &  & 0.6707 & 0.6463 &  & \textbf{0.6236} & \textbf{0.5154} \\
    & ToMe &  & \textbf{0.8073} & \textbf{0.8014} &  & \underline{0.6448} & \underline{0.6448} &  & 0.7331 & 0.4941 &  & \underline{0.7978} & \textbf{0.7641} &  & 0.5904 & 0.4210 \\
    & ToFU &  & \textbf{0.8073} & \underline{0.8010} &  & \textbf{0.6552} & \textbf{0.6552} &  & 0.7360 & 0.4881 &  & \textbf{0.7987} & \underline{0.7617} &  & 0.5863 & 0.4165 \\
    & DART &  & 0.7907 & 0.7876 &  & 0.6414 & 0.6414 &  & 0.7013 & 0.4419 &  & 0.7180 & 0.6745 &  & 0.4759 & 0.3440 \\
    & Ours & kh,x,0,x,10 & 0.7989 & 0.7974 & x,v,0,v,10 & 0.6138 & 0.6138 & q,q,10,q,10 & \textbf{0.8018} & \textbf{0.5527} & x,q,5,qh,10 & 0.7402 & 0.7143 & vh,qh,6,v,10 & \underline{0.6121} & \underline{0.4571} \\

    \midrule
    \multirow{6.5}{*}{TFM} 
    & Original\footnotemark[1] &  & 0.8208 & 0.8144 &  & 0.5862 & 0.5862 &  & 0.7696 & 0.5388 &  & 0.7798 & 0.7391 &  & 0.5703 & 0.4247 \\
    \cmidrule(lr){2-17}
    & EVIT &  & 0.8000 & 0.7979 &  & 0.5931 & 0.5931 &  & 0.7462 & 0.3643 &  & 0.7500 & 0.7149 &  & 0.6019 & 0.4402 \\
    & ToMe &  & 0.8070 & \underline{0.7994} &  & 0.6172 & 0.6172 &  & \textbf{0.7659} & 0.4615 &  & 0.7624 & 0.7256 &  & 0.5442 & 0.4280 \\
    & ToFU &  & \textbf{0.8080} & \textbf{0.7998} &  & 0.6138 & 0.6138 &  & 0.7651 & 0.4724 &  & 0.7655 & 0.7294 &  & 0.5377 & 0.4228 \\
    & DART &  & 0.7935 & 0.7893 &  & 0.5724 & 0.5724 &  & 0.7505 & 0.5060 &  & 0.7625 & 0.7226 &  & 0.4626 & 0.3486 \\
    & Ours & qh,qh,10,qh,10 & \underline{0.8076} & 0.7977 & q,v,1,x,10 & \textbf{0.6207} & \textbf{0.6207} & k,x,8,vh,10 & \textbf{0.7659} & \textbf{0.5399} & x,q,5,x,10 & \textbf{0.7818} & \textbf{0.7376} & qh,vh,10,kh,10 & \textbf{0.6206} & \textbf{0.4802} \\

    \bottomrule

\end{tabular}

\begin{tablenotes}
\footnotesize 
    \item Experiment results with \(r\), the number of tokens to be reduced per reduction, set to the biggest integer smaller than \(\frac{N}{l}\). \(N\) is the number of total tokens and \(l\) is set to 4 for BIOT, TFM and EEGPT and 12 for LaBraM. For BIOT and TFM, which consist of 4 layers of encoder, and LaBraM, which consists of 12 layers of encoder, we insert reductions in every layer. For EEGPT, which consists of 8 layer of encoder, reductions are insert at the 2nd, 4th, 6th and 8th layer. All hyper-parameters are set based on recommendations from original works. Compression ratio is defined as $(N - l\cdot r) / N$.
    \item[1] We include experiment results without token compression as Original.
\end{tablenotes}

\end{threeparttable}

    \caption{Performance comparison under the maximum available compression ratios.}
    \label{tab:4_model_1_ratio}
\end{table*}

\begin{table*}[ht]
    \centering
    \begin{threeparttable}
\begin{tabular}{cl cccccccc}
    \toprule
    \multirow{2.5}{*}{\textbf{Dataset}} & \multirow{2.5}{*}{\textbf{Method}} & \multicolumn{2}{c}{\textbf{20\%}} & \multicolumn{2}{c}{\textbf{40\%}} & \multicolumn{2}{c}{\textbf{60\%}} & \multicolumn{2}{c}{\textbf{80\%}} \\
    \cmidrule(lr){3-4} \cmidrule(lr){5-6} \cmidrule(lr){7-8} \cmidrule(lr){9-10}
    & &  \textbf{Acc} &  \textbf{BAcc} &  \textbf{Acc} &  \textbf{BAcc} &  \textbf{Acc} &  \textbf{BAcc} &  \textbf{Acc} &  \textbf{BAcc} \\

    \midrule
    \multirow{7}{*}{\makecell[c]{\textbf{TUAB} \\ (BIOT)}} 
    & Original\footnotemark[1] & 0.8081 & 0.8009 & 0.8081 & 0.8009 & 0.8081 & 0.8009 & 0.8081 & 0.8009 \\
    \cmidrule{2-10}
    & ToMe & 0.8090 & 0.8020 & 0.8087 & 0.8018 & 0.8086 & 0.8021 & 0.8082 & 0.8022 \\
    & ToFU & 0.8092 & 0.8022 & 0.8086 & 0.8017 & 0.8080 & 0.8014 & 0.8092 & 0.8029 \\
    & EViT & 0.8091 & \textbf{0.8028} & 0.8064 & 0.8011 & 0.8039 & 0.8000 & 0.7989 & 0.7966 \\
    & DART & 0.8095 & \textbf{0.8028} & 0.8085 & 0.8025 & 0.8052 & 0.8002 & 0.7979 & 0.7943 \\
    & Ours & \textbf{0.8093} & \underline{0.8026} & \textbf{0.8100} & \textbf{0.8036} & \textbf{0.8102} & \textbf{0.8045} & \textbf{0.8106} & \textbf{0.8061} \\    

    \midrule
    \midrule

    \multirow{2.5}{*}{\textbf{Dataset}} & \multirow{2.5}{*}{\textbf{Method}} & \multicolumn{2}{c}{\textbf{20\%}} & \multicolumn{2}{c}{\textbf{40\%}} & \multicolumn{2}{c}{\textbf{60\%}} & \multicolumn{2}{c}{\textbf{80\%}} \\
    \cmidrule(lr){3-4} \cmidrule(lr){5-6} \cmidrule(lr){7-8} \cmidrule(lr){9-10}
    & & \textbf{Acc} & \textbf{BAcc} & \textbf{Acc} & \textbf{BAcc} & \textbf{Acc} & \textbf{BAcc} & \textbf{Acc} & \textbf{BAcc} \\

    \midrule
    \multirow{7}{*}{\makecell[c]{\textbf{TUEV} \\ (LaBraM)}} 
    & Original\footnotemark[1]   & 0.8238 & 0.6585& 0.8238 & 0.6585& 0.8238 & 0.6585& 0.8238 & 0.6585 \\
    \cmidrule{2-10}
    & ToMe & 0.8185 & 0.6602 & 0.8180 & \textbf{0.6608} & 0.8064 & 0.6407 & 0.8071 & 0.6366 \\
    & ToFU & 0.8185 & 0.6603 & 0.8211 & \underline{0.6563} & 0.8100 & 0.6484 & 0.8144 & 0.6319 \\
    & EViT & 0.8243 & 0.6480 & 0.8227 & 0.6021 & 0.8206 & 0.5221 & 0.8155 & 0.4542 \\
    & DART & 0.8187 & 0.6531 & 0.8078 & 0.6355 & 0.7930 & 0.6179 & 0.7820 & 0.5704 \\   
    & Ours & \textbf{0.8356} & \textbf{0.6607} & \textbf{0.8340} & 0.6541 & \textbf{0.8283} & \textbf{0.6542} & \textbf{0.8237} & \textbf{0.6397} \\
   
\bottomrule
\end{tabular}

\begin{tablenotes}
    \footnotesize
    \item[1] We show results without token compression for easier comparison.
\end{tablenotes}

\end{threeparttable}

    \caption{Performance evaluation across varying token compression ratios.}
    \label{tab:2_model_4_ratio}
\end{table*}

\section{Limitations}

While \method effectively reduces the sequence length processed by subsequent Transformer layers, the compression module itself introduces minor computational overhead. In particular, the target token matching stage requires pairwise similarity computations between the selected source tokens and the retained tokens, leading to quadratic complexity relative to the participating tokens. Nevertheless, this overhead is substantially lower than that of full self-attention, as matching is restricted to a small subset of tokens after redundancy identification. Consequently, real-world inference latency remains significantly reduced. Exploring linear-complexity approximations for the matching stage while preserving compression also represents a promising direction for future research.

\end{document}